\documentclass[twoside]{article}

\usepackage[accepted]{aistats2022}

\usepackage{amsmath,amssymb,amsthm}
\usepackage{bm}
\usepackage{graphicx,float,wrapfig}
\usepackage{wrapfig}
\usepackage{ifpdf}
\usepackage{listings}
\usepackage{diagbox}
\usepackage{algorithm}
\usepackage{enumitem}
\usepackage{multirow}
\usepackage{multicol}
\usepackage{color}
\usepackage{soul}
\usepackage{hhline}
\usepackage{thm-restate}
\usepackage{graphicx}
\usepackage[font=small,labelfont=bf]{caption}
\usepackage{hyperref} 
\usepackage[round]{natbib}
\usepackage{aligned-overset}
\usepackage{subcaption}
\usepackage{pifont}
\usepackage{mathtools}
\usepackage{thm-restate}
\usepackage{tikz}
\usetikzlibrary{arrows.meta,calc,decorations.markings,math,arrows.meta}
\tikzset{every picture/.style={line width=1pt}} %

\newcommand*\colourcheck[1]{%
  \expandafter\newcommand\csname #1check\endcsname{\textcolor{#1}{\ding{52}}}%
}
\newcommand*\colourcross[1]{%
  \expandafter\newcommand\csname #1cross\endcsname{\textcolor{#1}{\ding{56}}}%
}
\colourcheck{red}
\colourcross{green}

\newcommand{\real}{\mathbb{R}}
\newcommand{\realp}{\Re}
\newcommand{\imag}{\Im}
\newcommand{\comp}{\mathbb{C}}

\newcommand{\bx}{\mathbf{x}}
\newcommand{\by}{\mathbf{y}}
\newcommand{\bA}{\mathbf{A}}
\newcommand{\bB}{\mathbf{B}}
\newcommand{\bC}{\mathbf{C}}
\newcommand{\bM}{\mathbf{M}}
\newcommand{\bV}{\mathbf{V}}
\newcommand{\bQ}{\mathbf{Q}}
\newcommand{\bU}{\mathbf{U}}
\newcommand{\bigo}{\mathcal{O}}

\newcommand{\bz}{\mathbf{z}}

\newcommand{\bv}{\mathbf{v}}
\newcommand{\bu}{\mathbf{u}}
\newcommand{\bw}{\mathbf{w}}
\newcommand{\zero}{\mathbf{0}}

\newcommand{\jacobian}{\mathbf{J}}
\newcommand{\iden}{\mathbf{I}}
\newcommand*{\hermconj}{^{\mathsf{H}}}
\newcommand{\bmat}[1]{\begin{bmatrix}#1\end{bmatrix}}
\newcommand{\bsmat}[1]{\left[ \begin{smallmatrix} #1 \end{smallmatrix} \right]}

\newtheorem{theorem}{Theorem}
\newtheorem{rem}{Remark}

\newtheorem{assumption}{Assumption} 

\declaretheorem[name=Corollary]{cor}

\newtheorem{definition}{Definition}

\hypersetup{
    colorlinks=true,
    linkcolor=blue,
    citecolor=blue,
    filecolor=blue,
    urlcolor=blue}

\begin{document}

\twocolumn[

\aistatstitle{Near-optimal Local Convergence of Alternating Gradient Descent-Ascent for Minimax Optimization}

\aistatsauthor{ Guodong Zhang \And Yuanhao Wang \And Laurent Lessard \And Roger Grosse }

\aistatsaddress{ University of Toronto \And Princeton University \And Northeastern University \And University of Toronto } ]

\begin{abstract}
Smooth minimax games often proceed by simultaneous or alternating gradient updates.
Although algorithms with alternating updates are commonly used in practice, the majority of existing theoretical analyses focus on simultaneous algorithms for convenience of analysis. In this paper, we study alternating gradient descent-ascent~(Alt-GDA) in minimax games and show that Alt-GDA is superior to its simultaneous counterpart~(Sim-GDA) in many settings. We prove that Alt-GDA achieves a near-optimal local convergence rate for strongly convex-strongly concave (SCSC) problems while Sim-GDA converges at a much slower rate.
To our knowledge, this is the \emph{first} result of any setting showing that Alt-GDA converges faster than Sim-GDA by more than a constant.
We further adapt the theory of integral quadratic constraints (IQC) and show that Alt-GDA attains the same rate \emph{globally} for a subclass of SCSC minimax problems.
Empirically, we demonstrate that alternating updates speed up GAN training significantly and the use of optimism only helps for simultaneous algorithms.
\end{abstract}

\begin{figure}[t]
\vspace{-0.1cm}
\begin{center}
    \begin{subfigure}{0.34\columnwidth}
        \includegraphics[width=\columnwidth]{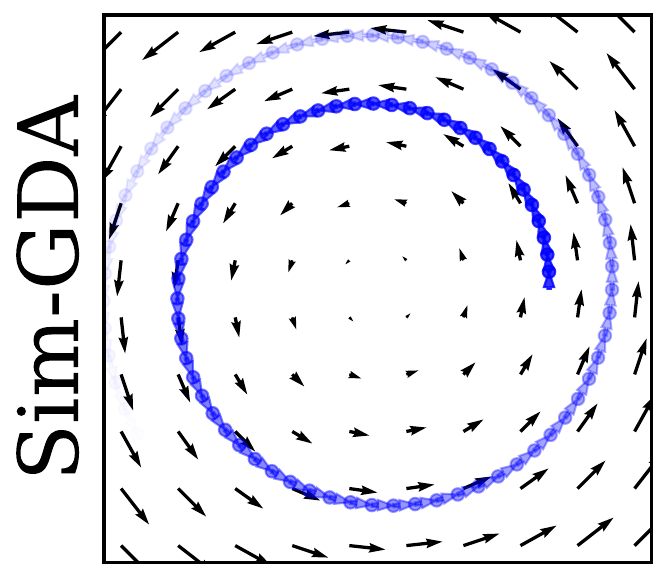}
    \end{subfigure}
    \begin{subfigure}{0.30\columnwidth}
        \includegraphics[width=\columnwidth]{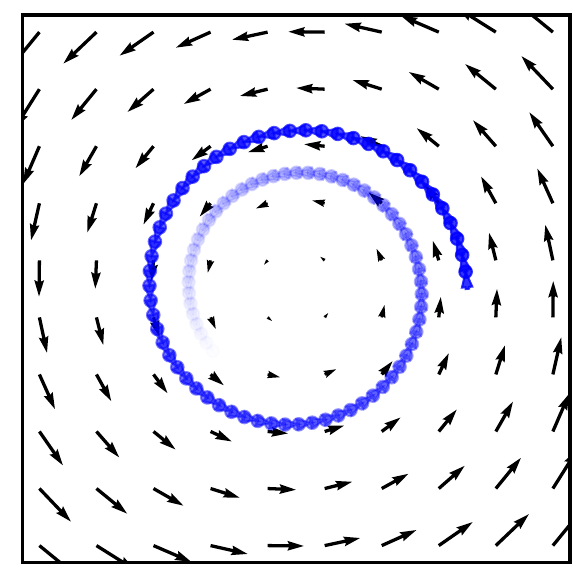}
    \end{subfigure}
    \begin{subfigure}{0.30\columnwidth}
        \includegraphics[width=\columnwidth]{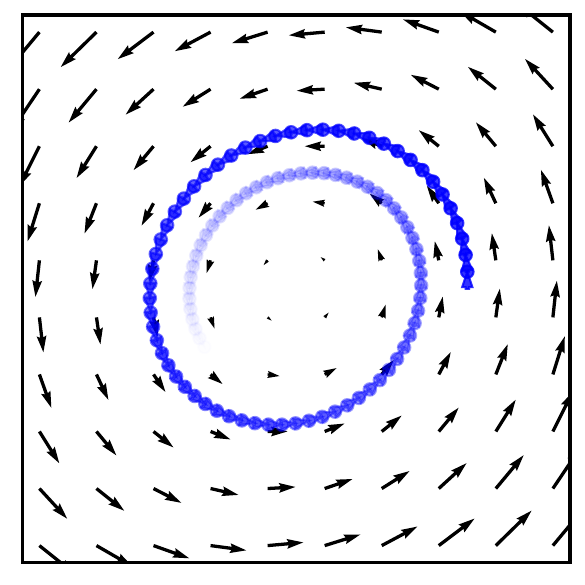}
    \end{subfigure}

    \begin{subfigure}{0.34\columnwidth}
        \includegraphics[width=\columnwidth]{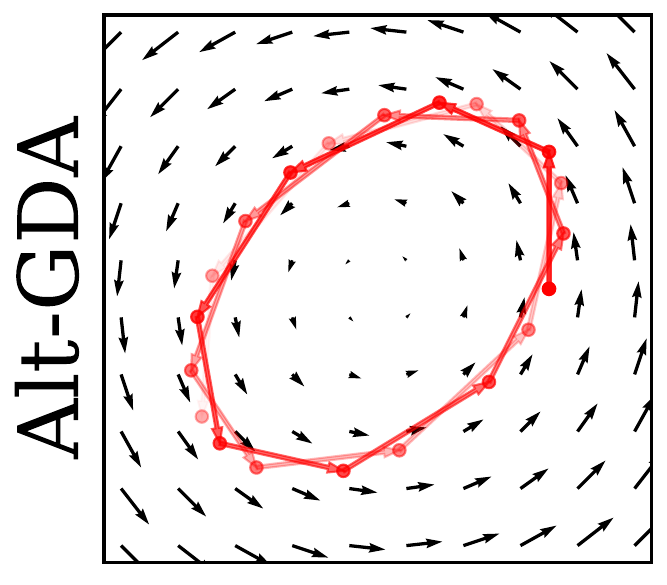}
    \end{subfigure}
    \begin{subfigure}{0.30\columnwidth}
        \includegraphics[width=\columnwidth]{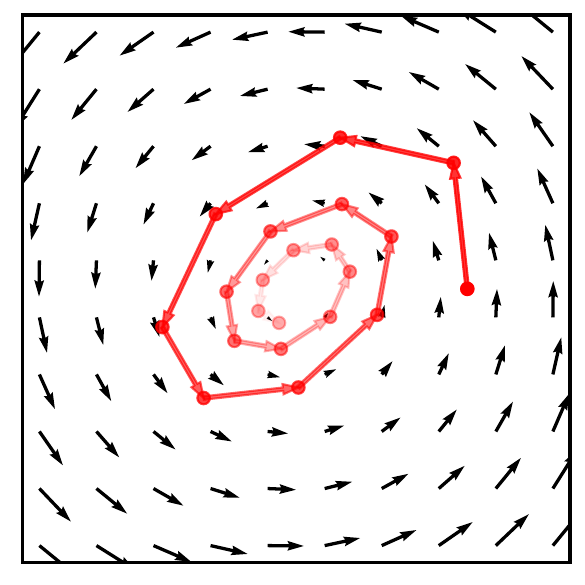}
    \end{subfigure}
    \begin{subfigure}{0.30\columnwidth}
        \includegraphics[width=\columnwidth]{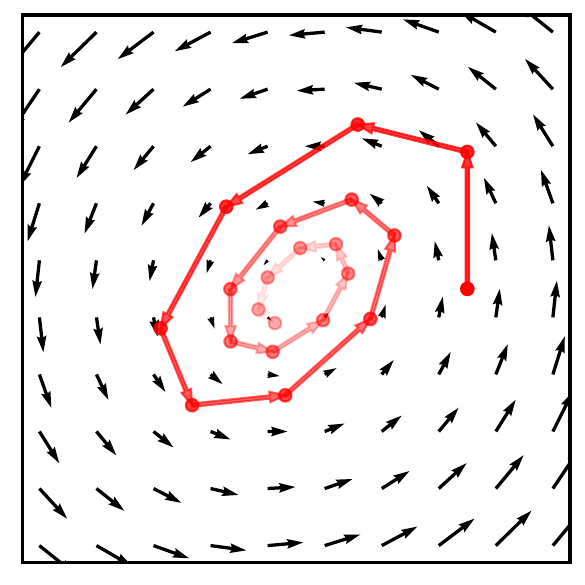}
    \end{subfigure}
\end{center}
    \vspace{-0.20cm}
	\caption{
	\textbf{Left:} a simple bilinear game with function $f(x, y)=10 xy$; \textbf{Middle:} a SCSC minimax game with $f(x, y)=0.5x^2 + 10xy - 0.5y^2$. \textbf{Right:} a minimax game that is not strongly-convex in $x$ with $f(x, y) = 10xy - y^2$;
	Color from dark to light indicates the direction of the trajectory.
	\textbf{Key Observation:} Alt-GDA converges much faster than Sim-GDA when they both converge.}
	\label{fig:fig1}
	\vspace{-0.2cm}
\end{figure}

\section{INTRODUCTION}
Since the seminal work of von Neumann~\citep{neumann1928theorie}, minimax optimization in the form of $\min_\bx \max_\by f(\bx, \by)$ has been a major focus of research in mathematics, economics and computer science~\citep{von1944theory, bacsar1998dynamic, roughgarden2010algorithmic}. Recently, minimax optimization has gained tremendous attention in machine learning as it offers a flexible paradigm that goes beyond ordinary loss function minimization. In particular, there is an increasing set of models that can be formulated as minimax problems, including (but not limited to) generative adversarial networks~\citep{goodfellow2014generative, arjovsky2017wasserstein}, adversarial training~\citep{madry2018towards}, robust optimization~\citep{ben2009robust} and primal-dual reinforcement learning~\citep{du2017stochastic, yang2020off}. 

The most natural and frequently used method for solving minimax problems is a generalization of gradient descent known as gradient descent-ascent (GDA), with either simultaneous or alternating updates of the two players, referred to as Sim-GDA and Alt-GDA, respectively, throughout the sequel. 
Unlike gradient descent, which converges to a local minimum for minimization problems under a broad range of conditions~\citep{lee2016gradient, lee2017first}, it is known that GDA with constant step-sizes can fail to converge for general smooth functions~\citep{mescheder2017numerics}, even for unconstrained bilinear games~\citep{gidel2019negative, bailey2018multiplicative}. Even when it does converge, GDA may exhibit rotational behaviors~\citep{mescheder2017numerics, letcher2019differentiable, florian2019competitive} and hence converge slowly (see Figure~\ref{fig:fig1}). 
To combat these issues, several algorithms have been introduced specifically for smooth minimax games, including consensus optimization~\citep{mescheder2017numerics}, symplectic gradient adjustment~\citep{letcher2019differentiable}, negative momentum (NM)~\citep{gidel2019negative, zhang2020suboptimality}, optimistic gradient descent-ascent (OGDA)~\citep{popov1980modification, rakhlin2013optimization, daskalakis2018training, mertikopoulos2018optimistic} and extra-gradient (EG)~\citep{korpelevich1976extragradient}. 

In theory, many of these algorithms enjoy improved convergence rates compared to GDA. In particular, both OGDA and EG are near-optimal for SCSC minimax problems~\citep{mokhtari2020convergence}.
However, in practice, GDA and its adaptive variants are still the go-to algorithms for many applications (e.g., GAN optimization and offline policy evaluation~\citep{yang2020off}). 
Here, the catch is that the overwhelming majority of existing theoretical analyses focus on simultaneous algorithms where players update their strategies at the same time, as simultaneous updates are easier to analyze and can often be formulated as solving a variational inequality problem~\citep{harker1990finite, gidel2018variational, zhang2020unified}.
This is in stark contrast to our common practice where alternating algorithms are actually used.
Nonetheless, our understanding of alternating algorithms in minimax optimization is severely limited to simple bilinear games.
Despite it being a very natural question to ask, the convergence properties of Alt-GDA for SCSC minimax games and many other settings remain largely unknown.
The key difficulty is that every iteration of an alternating algorithm is a composition of two half updates, which greatly complicates analysis.

\textbf{Our contributions.}
In this paper, we take a step towards understanding Alt-GDA and closing the gap between theory and practice.
We first revisit the convergence properties of Alt-GDA in bilinear games for completeness. 
We then discuss our main contributions on proving near-optimal convergence rates of Alt-GDA.
In more detail:
\vspace{-0.25cm}
\begin{enumerate}[leftmargin=2em]\setlength\itemsep{0.01em}
    \item We prove that, for SCSC minimax games\footnote{The SCSC setting is fundamental. Via reduction~\citep{lin2020near, yang2020catalyst}, an efficient algorithm for this setting implies efficient algorithms for other settings, including strongly convex-concave, convex-concave, and non-convex-concave settings.}, Alt-GDA achieves an iteration complexity of $\bigo(\kappa)$ locally ($\kappa$ is the condition number), which is quadratically better than the $\bigo(\kappa^2)$ bound for Sim-GDA and even matches EG/OGDA. 
    Importantly, the complexity bound for Alt-GDA in this setting is near-optimal as it matches the coarse lower bound in~\cite[Corollary 1]{azizian2020accelerating}.
    \item We further prove that both Sim-GDA and Alt-GDA attain linear convergence when the minimax problem has only strong concavity in $\by$ but no strong convexity in $\bx$ by assuming non-singularity of the coupling matrix. 
    \item We show that Alt-GDA can converge with the same rate $\bigo(\kappa)$ globally for a class of SCSC minimax games with a bilinear coupling term. This is done by using theory of IQC to automatically search for a Lyapunov function.
    \item Lastly, we validate our theory on quadratic minimax games. Empirically, we demonstrate that alternating updates could speed up GAN training dramatically (which matches the existing results in \citet{goodfellow2014generative, radford2015unsupervised}) and perform on par with optimistic updates though GAN objective is generally nonconvex-nonconcave.
\end{enumerate}

\section{PRELIMINARIES}
\textbf{Notation}. In this paper, scalars are denoted by lower-case letters (e.g., $\lambda$), vectors by lower-case bold letters (e.g., $\bx$), matrices by upper-case bold letters (e.g., $\jacobian$). 
The spectrum of a square matrix $\bA$ is denoted by $\text{Sp}(\bA)$, and a generic eigenvalue by $\lambda$. 
We respectively note $\sigma_\text{min}(\bA)$ and $\sigma_\text{max}(\bA)$ the smallest and the largest positive singular values of $\bA$. For matrix inequality $\bA \succeq \bB$, we mean $\bA - \bB$ is positive semi-definite.
We use $\realp$ and $\imag$ to denote the real part and imaginary part of a complex scalar respectively. 
We use $\real$ and $\comp$ to denote the set of real numbers and complex numbers, respectively.
We use $\rho(\bA) = \lim_{t \rightarrow \infty} \|\bA^t\|^{1/t}$ to denote the spectral radius of matrix $\bA$. $\bigo$, $\Omega$ and $\Theta$ are standard asymptotic notations. 

\vspace{-0.1cm}
\subsection{Two-player Minimax Games}
\vspace{-0.1cm}
We begin by presenting the fundamental two-player zero-sum game that we will consider in the sequel. To be specific, our problem of interest is the following unconstrained minimax optimization problem:
\begin{equation}
    \min_{\bx \in \real^m} \max_{\by \in \real^n} f(\bx, \by).
    \vspace{-0.1cm}
\end{equation}
We are usually interested in finding a \emph{Nash equilibrium}~\citep{von1944theory}: a set of parameters from which no player can (unilaterally) improve its objective function.
In this work, we focus on the case of $f$ being a convex-concave and smooth function. Here we state the assumption formally.
\begin{assumption}
The function $f$ is continuously differentiable and $L$-smooth in $\bx$ and $\by$. Furthermore, we assume $f$ is convex in $\bx$ and concave in $\by$.
\end{assumption}
For completeness, we state the definition of smooth function. We note that the smoothness assumption is standard for convergence analysis in the literature.
\begin{definition}
A differentiable function $\phi: \real^d \rightarrow \real$ is $L$-smooth if it has $L$-Lipschitz gradient on $\real^d$, i.e., for any $\bx_1, \bx_2 \in \real^d$, we have $\|\nabla\phi(\bx_1) - \nabla \phi(\bx_2)\| \leq L \|\bx_1 - \bx_2\|$.
\end{definition}
One of the nice properties of working with convex-concave problems is that there often exists at least one \emph{global} Nash equilibrium $(\bx^*, \by^*)$ such that for any $\bx \in \real^m, \by \in \real^n$ we have 
\begin{equation*}
    f(\bx^*, \by) \leq f(\bx^*, \by^*) \leq f(\bx, \by^*).
\end{equation*}

\subsection{Gradient Descent-Ascent Family}
We now present two algorithms we will discuss in this paper, Sim-GDA and Alt-GDA.
The de-facto standard algorithm for finding Nash equilibria of general smooth two-player minimax games
is simultaneous gradient descent-ascent (Sim-GDA) which is a direct generalization of gradient descent to minimax games. In particular, it updates both players $\bx$ and $\by$ simultaneously:
\begin{equation}\label{eq:sim-gda-update}
\begin{aligned}
     \bx_{t+1} &= \bx_t - \eta \nabla_\bx f(\bx_t, \by_t), \\
     \by_{t+1} &= \by_t + \eta \nabla_\by f(\bx_t, \by_t),
\end{aligned}
\end{equation}
where $\eta$ is the step size\footnote{Using separate step sizes for two players does not improve the worst-case convergence rate.}. 
Succinctly, Sim-GDA updates \eqref{eq:sim-gda-update}  can be defined as the repeated application of a nonlinear operator in the form of $\bz_{t+1} = F_\eta^\textup{Sim}(\bz_t) \triangleq \bz_t - \eta V(\bz_t)$ with $\bz = [\bx^\top, \by^\top]^\top$ and $V(\bz) = [\nabla_\bx f(\bx, \by)^\top, -\nabla_\by f(\bx, \by)^\top]^\top$, the gradient vector field.
By contrast, Alt-GDA takes advantage of the fact that the iterates $\bx_{t+1}$ and $\by_{t+1}$ are computed sequentially:
\vspace{-0.1cm}
\begin{equation}\label{eq:alt-gda-update}
\begin{aligned}
    \bx_{t+1} &= \bx_t - \eta \nabla_\bx f(\bx_t, \by_t), \\
    \by_{t+1} &= \by_t + \eta \nabla_\by f({\color{red}\bx_{t+1}}, \by_t).
\end{aligned}
\end{equation}
Similarly, we can write the updates as $\bz_{t+1} = F_\eta^\textup{Alt}(\bz_t)$.

\vspace{-0.1cm}
\subsection{Local Convergence Rates}
\vspace{-0.1cm}
We stress that local convergence analysis has been widely adopted in smooth game optimization (see e.g.,~\cite{gidel2019negative, wang2019solving, azizian2020accelerating, zhang2020suboptimality, liang2019interaction, fiez2020gradient}). Under certain conditions on a fixed point operator $F$, linear convergence is guaranteed in a neighborhood around a fixed point $\bz^*$ (i.e. local convergence).
\begin{theorem}[{\cite[Proposition~4.4.1]{bertsekas1997nonlinear}}]\label{thm:local}
    For a continuously differentiable nonlinear operator $F$ with the fixed point $\bz^*$, if the spectral radius $\rho_F \triangleq \rho(\nabla F(\bz^*)) < 1$, then for any $\bz_0$ in a neighborhood of $\bz^*$, the iterates of $\bz_t$ converge to $\bz^*$ with a linear rate of $\bigo((\rho_F + \epsilon)^t)$ for any $\epsilon >0$.
\end{theorem}
With this theorem, one can obtain local convergence rate of an algorithm by just computing the spectral radius of $\nabla F_\eta(\bz^*)$, which is a constant matrix depending on $\eta$ in our setting. In the paper, we focus on the \emph{worst-case} convergence rate which is defined (up to a $\epsilon$ difference) as follows:
\begin{equation}\label{eq:rate}
    \min_\eta \max_{F \in \mathcal{M}} \rho(\nabla F_\eta(\bz^*)),
    \vspace{-0.1cm}
\end{equation}
where the inner maximization is over all possible instances within the whole problem class $\mathcal{M}$.

\subsection{Revisiting Alt-GDA for Bilinear Games}
In this section, we revisit the unconstrained bilinear games~\citep{gidel2019negative, daskalakis2018last, liang2019interaction, mokhtari2020unified} for which Sim-GDA diverges with any finite step size. Formally, the bilinear game is given by
\begin{equation}
    \min_{\bx \in \real^m} \max_{\by \in \real^n} \bx^\top \bB \by,
    \vspace{-0.1cm}
\end{equation}
where we ignore the linear terms without loss of generality.
Here, the Nash equilibrium is $(\bx^*, \by^*)$ satisfying $\bB^\top \bx^* = \zero$ and $\bB\by^* = \zero$.
To measure convergence, one could monitor the distance to the equilibrium:
\begin{equation}
    \Delta_t = \|\bx_t - \bx^*\|_2^2 + \|\by_t - \by^*\|_2^2.
    \vspace{-0.1cm}
\end{equation}
We aim to understand the difference between the dynamics of simultaneous and alternating methods.
Practitioners have been widely using the latter instead of the former when optimizing GANs despite the rich optimization literature on simultaneous methods.

For Sim-GDA, the eigenvalues of $\iden - \nabla F_\eta^\textup{Sim}$ are all pure imaginary. As a result, we have the spectral radius as $\rho(\nabla F_\eta^\textup{Sim}) = 1 + \eta^2\sigma_\text{max}^2(\bB)$. Therefore, we have
\begin{theorem}[\cite{gidel2019negative}]
    For any $\eta > 0$, the iterates of Sim-GDA diverges as
    \begin{equation*}
        \Delta_t \in \Omega\left(\Delta_0 (1 + \eta^2 \sigma_\text{max}^2(\bB))^t \right)
    \end{equation*}
\end{theorem}
\vspace{-0.3cm}
This theorem states that the iterates of Sim-GDA diverge linearly for any positive constant step-size $\eta$. By contrast, the iterates of Alt-GDA stay bounded due to the sequential update rule which significantly shifts the eigenvalues of the Jacobian. Specifically, the eigenvalues of $\nabla F_\eta^\textup{Alt}$ are roots of the polynomial $ (x - 1)^2 + \eta^2 \lambda x$ with $\lambda \in \text{Sp}(\bB^\top \bB)$.
As a consequence, the spectral radius of $\nabla F_\eta^\textup{Alt}$ is upper bounded by $1$ for some $\eta$ and hence the iterates of Alt-GDA stays bounded. 
\begin{theorem}\label{thm:alt-gda-bilinear}
For any $0 < \eta \leq \frac{2}{\sigma_\text{max}(\bB)}$, the iterates of Alt-GDA stay bounded
\begin{equation*}
    \Delta_t \in \bigo\left(\Delta_0 \right)
\end{equation*}
\end{theorem}
\vspace{-0.2cm}
Similar results can be found in the literature (see e.g., \cite{gidel2019negative, zhang2019convergence}). In addition, one can show that for bilinear games, Alt-GDA is a symplectic integrator applied on the continuous dynamics~\citep{bailey2020finite}, which preserves energy and volume.

\vspace{-0.15cm}
\section{NEAR-OPTIMAL LOCAL CONVERGENCE IN SCSC SETTING}\label{sec:sec-4}
\vspace{-0.15cm}
Bilinear games, as discussed previously, are somewhat simplistic in that they obey a conservation law and can be easily solved by performing gradient descent on the Hamiltonian~\citep{letcher2019differentiable, azizian2020accelerating}. In this section, we consider a different class of games whose Jacobian has both symmetric and antisymmetric components, and are therefore arguably harder to solve.
In particular, we assume $f(\bx, \by)$ is SCSC and smooth, which implies
\begin{equation*}
\begin{aligned}
    \mu_\bx \iden \preceq \nabla_\bx^2 f \preceq L_\bx \iden, &\; \mu_\by \iden \preceq -\nabla_\by^2 f \preceq L_\by \iden, \\
    \|\nabla_{\bx\by}^2 f\|_2 &\leq L_{\bx\by}.
\end{aligned}
\end{equation*}
We let $L \triangleq \max\{L_\bx, L_\by, L_{\bx\by}\}$ and $\mu \triangleq \min\{\mu_\bx, \mu_\by\}$ and define the condition number $\kappa \triangleq L/\mu$. Accordingly, one can define $\kappa_\bx \triangleq L / \mu_\bx$ and $\kappa_\by \triangleq L / \mu_\by$. 
We now briefly summarize some known results about convergence of Sim-GDA in this setting. The worst-case convergence rate \eqref{eq:rate} of Sim-GDA reduces to $\rho(\iden - \eta\nabla V(\bz^*))$, which is equivalent to 
\vspace{-0.1cm}
\begin{equation}\label{eq:rate-quad}
    \min_{\eta} \max_{\lambda \in \mathcal{K}} |1 - \eta \lambda |,
    \vspace{-0.1cm}
\end{equation}
where $\mathcal{K}$ is the support of the eigenvalues of the Jacobian of the gradient vector field $V$. It can be shown that $\mathcal{K} = \left\{ \lambda \in \comp: |\lambda| \leq \sqrt{2}L, \realp \lambda \geq \mu > 0 \right\}$ (see Appendix~\ref{app:B}). This set is the intersection between a circle and a halfplane~\citep{azizian2020accelerating}. 
Eqn.~\eqref{eq:rate-quad} leaves open the choice of $\eta$, and it is known that the presence of large imaginary eigenvalues of the Jacobian forces a small value of $\eta$, thereby limiting the rate of convergence~\citep{mescheder2017numerics}. We summarize the result below:
\begin{theorem}\label{thm:sim-gda}
    With the step size $\eta = \frac{\mu}{2L^2}$, we have $\rho(\nabla F_\eta^\textup{Sim}(\bz^*)) < 1 - \frac{1}{ 4\kappa^2}$. Hence, Sim-GDA converges locally at a linear rate $\bigo\left(\left(1 - \frac{1}{4\kappa^2}\right)^t\right)$.
\end{theorem}
This theorem suggests that Sim-GDA converges to the equilibrium linearly with an iteration complexity of $\bigo(\kappa^2)$, which is known to be tight~\citep{azizian2020accelerating} but much slower than the $\bigo(\kappa)$ iteration complexity of extra-gradient (EG) or optimistic gradient-descent-ascent (OGDA)~\citep{gidel2018variational, mokhtari2020unified, azizian2020tight, zhang2020unified}.

\begin{figure}[t]
	\centering
    \includegraphics[width=0.7\columnwidth]{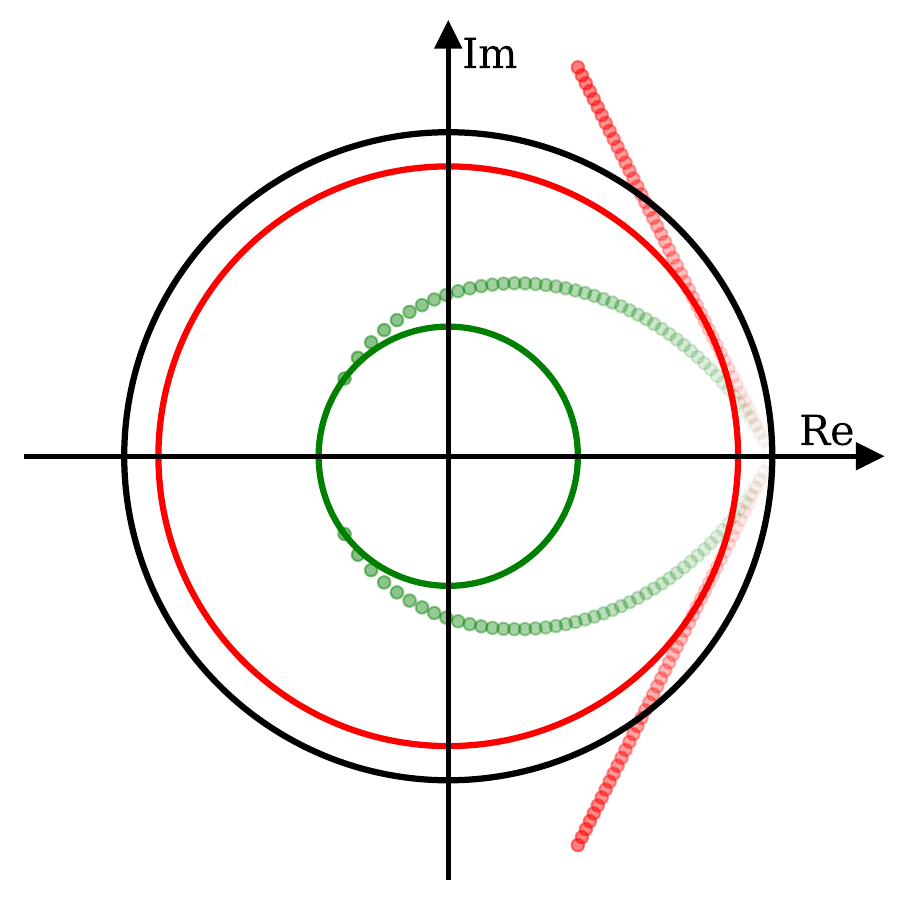}
    \vspace{-0.1cm}
    \caption{Eigenvalues of $\nabla F_\eta^\textup{Sim}$ and $\nabla F_\eta^\textup{Alt}$ for a minimax problem with the function $f(x, y) = 0.3 x^2 + 1.2 x  y - 0.3 y^2$. For a fixed step-size $\eta$, eigenvalues of Sim-GDA are represented with {\color{red}\textbf{red dots}} and eigenvalues of Alt-GDA are {\color[rgb]{0,0.6,0}\textbf{green dots}}. Their trajectories as $\eta$ sweeps in $[0, 1]$ are shown from light colors to dark colors. Convergence circles for Sim-GDA are in {\color{red}\textbf{red}}, Alt-GDA in {\color[rgb]{0,0.6,0}\textbf{green}}, and unit circle in \textbf{black}. The convergence circles are optimized over all step-sizes. Alternating updates help as its convergence circle (green) is smaller, due to the fact that it allows us to use much larger step-sizes. Figure inspired by~\cite{gidel2019negative}.}
	\label{fig:sim-alt-comp}
	\vspace{-0.25cm}
\end{figure}
To understand why, we note the maximization over $\lambda$ in \eqref{eq:rate-quad} is attained by %
$\lambda = \mu + \sqrt{2L^2 - \mu^2}i$, which has a large imaginary component. 
It is easy to show (see e.g.,~\cite{mescheder2017numerics}) that the largest feasible step size in \eqref{eq:rate-quad} is inversely proportional to $(|\lambda|/\realp(\lambda))^2$. Hence, the step size has to be extremely small in the presence of eigenvalues with large imaginary parts, which in turn, leads to slow convergence.
In a nutshell, the culprits of slow convergence in Sim-GDA are eigenvalues of the Jacobian of the associated vector field $V$ with large imaginary parts. We stress that eigenvalues with large imaginary components contribute to a strong ``rotational force''. To improve convergence, many algorithms have been introduced to suppress the rotational force, including EG, OGDA, and NM. 
Indeed, all three of these algorithms improve the convergence rate by some margin in theory. Nevertheless, GDA (or its adaptive variant) is still the go-to algorithm in practice.

We believe the reason these alternative algorithms haven't been adopted widely is that practical algorithms for cases such as GANs are typically based on Alt-GDA rather than Sim-GDA. Surprisingly, despite the popularity of Alt-GDA, its convergence properties haven't been analyzed in this setting.
While it is perhaps intuitive that Alt-GDA should perform better than Sim-GDA due to its use of fresher gradient information, we show that, in fact, Alt-GDA achieves a quadratic speedup over Sim-GDA locally and matches the convergence rate of EG and OGDA. 

Local convergence analyses of Sim-GDA, EG and OGDA are based on matrix spectral calculation, and in principle one can apply this to Alt-GDA as well. However, bounding the spectral radius is much harder for Alt-GDA, since the algorithm involves two half steps, and the spectral radius of the matrix product can't be bounded straightforwardly in terms of the spectral radii of the two factors. This is likely why the convergence rate of Alt-GDA remained unknown. By treating complex eigenvalues differently and adopting a fined-grained analysis, we arrive at the following bounds for the eigenvalues of $\nabla F^\textup{Alt}_\eta(\bz^*)$:
\begin{theorem}\label{thm:alt-gda}
    With the step size $\eta \leq \frac{1}{2L}$, the eigenvalues of $\nabla F^\textup{Alt}_\eta(\bz^*)$ satisfy
    \vspace{-0.1cm}
    \begin{equation*}
    \begin{aligned}
        \text{if real: } & |\lambda| \leq \max\{1 - \eta \mu_\bx, 1 -  \eta \mu_\by \}, \\
        \text{if complex: } & |\lambda| \leq \sqrt{(1 - \eta \mu_\bx) (1 - \eta \mu_\by)}.
    \end{aligned}
    \end{equation*}
\end{theorem}
\vspace{-0.1cm}
\begin{rem}
    In stark contrast to Sim-GDA, for which the complex eigenvalues of $\nabla F_\eta^\textup{Sim}$ can have magnitude as large as $\sqrt{1 - 2\eta \mu + 2\eta^2 L^2}$, the complex eigenvalues of $\nabla F_\eta^\textup{Alt}$ are much smaller in magnitude and are even smaller than the real eigenvalues as shown in Theorem~\ref{thm:alt-gda}.
    As a result, we are allowed to use a larger step size, which gives an improved convergence rate (see Figure~\ref{fig:sim-alt-comp} for details).
\end{rem}
Following immediately from Theorem~\ref{thm:alt-gda}, we have the following Corollary.
\begin{cor}\label{cor:alt-gda}
With $\eta = \frac{1}{2L}$, we have $\rho(\nabla F_\eta^\textup{Alt}(\bz^*)) \leq 1 - \frac{1}{2 \kappa}$.
Hence by Theorem~\ref{thm:local}, Alt-GDA converges locally at a linear rate $\bigo \left(\left(1 - \frac{1}{2\kappa} + \epsilon\right)^t\right)$ with $\epsilon > 0$ an arbitrarily small constant.
\end{cor}
In particular, this corollary suggests that the iteration complexity of Alt-GDA matches the coarse lower iteration complexity bound\footnote{The fine-grained bound~\citep{zhang2019lower} is $\Omega(\sqrt{\kappa_\bx \kappa_\by})$. One could achieve this bound by using a accelerated proximal point framework~\citep{yang2020catalyst} with Alt-GDA in the inner-loop.} $\Omega(\kappa)$~\citep[Corollary 1]{azizian2020accelerating} up to a constant, implying Alt-GDA is near-optimal (at least locally).
This is the \emph{first} time that one can rigorously show the Alt-GDA converges faster than Sim-GDA by more than a constant, let alone quadratically faster.

Furthermore, it implies that the convergence rate of Alt-GDA is no worse than its rate for pure cooperative games with $\bB \triangleq \nabla_{\bx\by}^2 f = \zero$. Put differently, the adversarial component (the existence of coupling matrix $\bB$) does \emph{not} make the optimization any harder for Alt-GDA. We remark that this is \emph{not} true for Sim-GDA because in that case, the coupling matrix $\bB$ introduces complex eigenvalues with large imaginary parts, which slow down convergence.

\section{ACCELERATION WITHOUT STRONG CONVEXITY}\label{sec:sec-5}
We have shown that Alt-GDA achieves a near-optimal local convergence rate for SCSC minimax games.
In this section, we further consider the case that has only strong concavity in the player $\by$ but \emph{no} strong convexity in $\bx$. In particular, it is equivalent to assuming
\begin{equation*}
\begin{aligned}
    \zero \preceq \nabla_\bx^2 f \preceq L_\bx \iden, \; &\mu_\by \iden \preceq -\nabla_\by^2 f \preceq L_\by \iden, \\ \|\nabla_{\bx\by}^2 f\|_2 &\leq L_{\bx\by}.
\end{aligned}
\end{equation*}
This setting was investigated in empirical policy evaluation where no strong convex regularization is applied on the primal variables~\citep{du2017stochastic}. They showed that the non-singularity of the coupling matrix $\bB \triangleq \nabla_{\bx\by}^2 f(\bx^*, \by^*)$ can help achieve linear convergence for Sim-GDA. Technically, the coupling matrix $\bB$ has to be full-row rank (i.e., $\lambda_\text{min}(\bB\bB^\top) > 0$) and we simply assume $\mu_{\bx\by} \triangleq \sigma_\text{min}(\bB) > 0$. Then for Sim-GDA, we have the eigenvalues of its Jacobian as follows:
\begin{theorem}\label{prop:sim-gda}
Let $\eta \leq \frac{1}{L}$, the eigenvalues of $\nabla F^\textup{Sim}_\eta(\bz^*)$ satisfy the following bound
\begin{equation*}
\begin{aligned}
   \text{if real: }& |\lambda| \leq \max\left\{1 - \tfrac{\eta}{L} \mu_{\bx\by}^2, 1- \eta \mu_\by \right\}, \\ \text{if complex: }& |\lambda| \leq \sqrt{1 - \eta \mu_\by + 2\eta^2 L^2}.
\end{aligned}
\end{equation*}
\end{theorem}
\vspace{-0.2cm}
To be noted, our eigenvalue bounds in Theorem~\ref{prop:sim-gda} are slightly different from that in \cite{du2017stochastic} as they allow step size separation for player $\bx$ and $\by$.
As a result, we get the following local convergence rate by optimizing over the step-size $\eta$.
\begin{cor}\label{thm:sim-gda-no}
    With $\eta = \frac{\mu_\by}{4 L^2}$, we have $\rho(\nabla F_\eta^\textup{Sim}(\bz^*)) < 1 - \frac{1}{16 \max\{\kappa_\by \kappa_{\bx\by}^2, \kappa_\by^2\}}$.
    Hence, Sim-GDA converges locally at a linear rate $\bigo \left(\left(1 - \frac{1}{16 \max\{\kappa_\by \kappa_{\bx\by}^2, \kappa_\by^2\}}\right)^t\right)$.
\end{cor}
This corollary suggests that the convergence rate of Sim-GDA could match the rate in Theorem~\ref{thm:sim-gda} if the coupling matrix is well-conditioned (i.e., $\kappa_{\bx\by} \approx 1$), albeit the absence of strong convexity in $\bx$.
Naturally, this begs the question: whether we can derive similar results for Alt-GDA that improves upon the rate bound of Sim-GDA.
We answer this question in the affirmative.
In particular, we have the following bounds for the eigenvalues of $\nabla F^\textup{Alt}_\eta$.
\begin{theorem}\label{prop:alt-gda}
Let $\eta \leq \frac{1}{2L}$, the eigenvalues of $\nabla F^\textup{Alt}_\eta(\bz^*)$ satisfy the following bound
\begin{equation*}
\begin{aligned}
    \text{if real: }& |\lambda| \leq \max\{1 - \eta^2 \mu_{\bx\by}^2, 1 - \eta \mu_\by \}, \\ 
    \text{if complex: }& |\lambda| \leq \sqrt{1 - \eta \mu_\by}.
\end{aligned}
\end{equation*}
\end{theorem}
\vspace{-0.2cm}
Compare the eigenvalue bound of Alt-GDA to Sim-GDA, one may notice that the main difference is the complex eigenvalues. Similar to the SCSC setting, the complex eigenvalues of Alt-GDA are much smaller in magnitude, thus allowing us to use larger step sizes.
Consequently, we have a better convergence rate for Alt-GDA (see Figure~\ref{fig:fig1} for detailed comparisons).
\begin{cor}
    Choosing $\eta = \frac{1}{2L}$, we have we have $\rho(\nabla F_\eta^\textup{Alt}(\bz^*)) \leq 1 - \frac{1}{4\max\{\kappa_{\bx\by}^2, \kappa_\by\}}$.
    Hence, Alt-GDA converges locally at a linear rate $\bigo((1 - \frac{1}{4\max\{\kappa_{\bx\by}^2, \kappa_\by\}} + \epsilon)^t)$ with $\epsilon > 0$ a small constant.
\end{cor}
Compared with the bound of Sim-GDA in Theorem~\ref{thm:sim-gda-no}, one can see that Alt-GDA converges much faster than Sim-GDA, especially when $\kappa_\by$ is large.

\section{GLOBAL CONVERGENCE FOR BILINEARLY-COUPLED MINIMAX GAMES}
So far, we derived local convergence rates of Alt-GDA in different settings. Nonetheless, the global convergence results remain largely unknown. Unlike local convergence analysis, we have to switch to Lyapunov theory for global convergence analysis. Finding a right Lyapunov function for Alt-GDA turns out to be extremely hard and we resort to integral quadratic constraints (IQC) theory~\citep{lessard2016analysis, zhang2020unified} for a computer-aided proof\footnote{See the blog by Adrien Taylor for more details about computer-aided analyses  (\url{https://francisbach.com/computer-aided-analyses/}).}. Basically, we view the algorithm as an interconnected dynamical system with nonlinear feedback (i.e., the gradient) and model the nonlinear feedback with quadratic constraints\footnote{Both convexity and smoothness can be characterized tightly with quadratic constraints.}. Then it allows us to \emph{automatically} search for a quadratic Lyapunov function for certifying the worst-case convergence rate by solving a semi-definite program (SDP). Due to space constraints, we refer the reader to Appendix~\ref{app:iqc} for all the details.

In particular, we analyze Alt-GDA for bilinearly-coupled minimax games with the following form:
\begin{equation}\label{eq:bilinear-sp}
    \min_{\bx \in \real^m} \max_{\by \in \real^n} f(\bx) + \bx^\top \bB \by - g(\by),
\end{equation}
where we assume both $f$ and $g$ are $\mu$-strongly-convex and $L$-smooth, $\|\bB\|_2 \leq L$. This problem is a special case of the minimax games that is amenable to IQC analysis. This problem has been studied extensively~\cite{chambolle2011first, du2019linear, xie2021dippa}. However, the convergence properties of Alt-GDA again remain unknown.

Using the IQC framework, we are able to search for the best possible convergence rate of Alt-GDA for every given condition number $\kappa$ by solving a SDP.
However, the size of the SDP is proportional to $m$ and $n$. 
This can be problematic in cases where $m$ (or $n$) is large because it can be computationally costly to solve large SDPs.
Fortunately, we prove that the high dimensional problem isn't any harder than than the case of $m = n = 1$, so we can reduce the problem to a SDP with $m = n = 1$, which is easy to solve.
\begin{restatable}{theorem}{iqc}\label{thm:sdp-reduction}
Using the IQC framework to analyze the convergence rate of Alt-GDA on problem \eqref{eq:bilinear-sp}, we can simply assume $m = n = 1$ if  $\bB$ is diagonal. Let $\rho_{m, n}$ be the IQC-certified rate for problem \eqref{eq:bilinear-sp} with $\bx \in \real^m$ and $\by \in \real^n$, then we have $\rho_{m,n} \leq \rho_{1, 1}$.
\end{restatable}

\begin{figure}[t]
\vspace{-0.15cm}
\centering
    \centering
    \includegraphics[width=0.85\columnwidth]{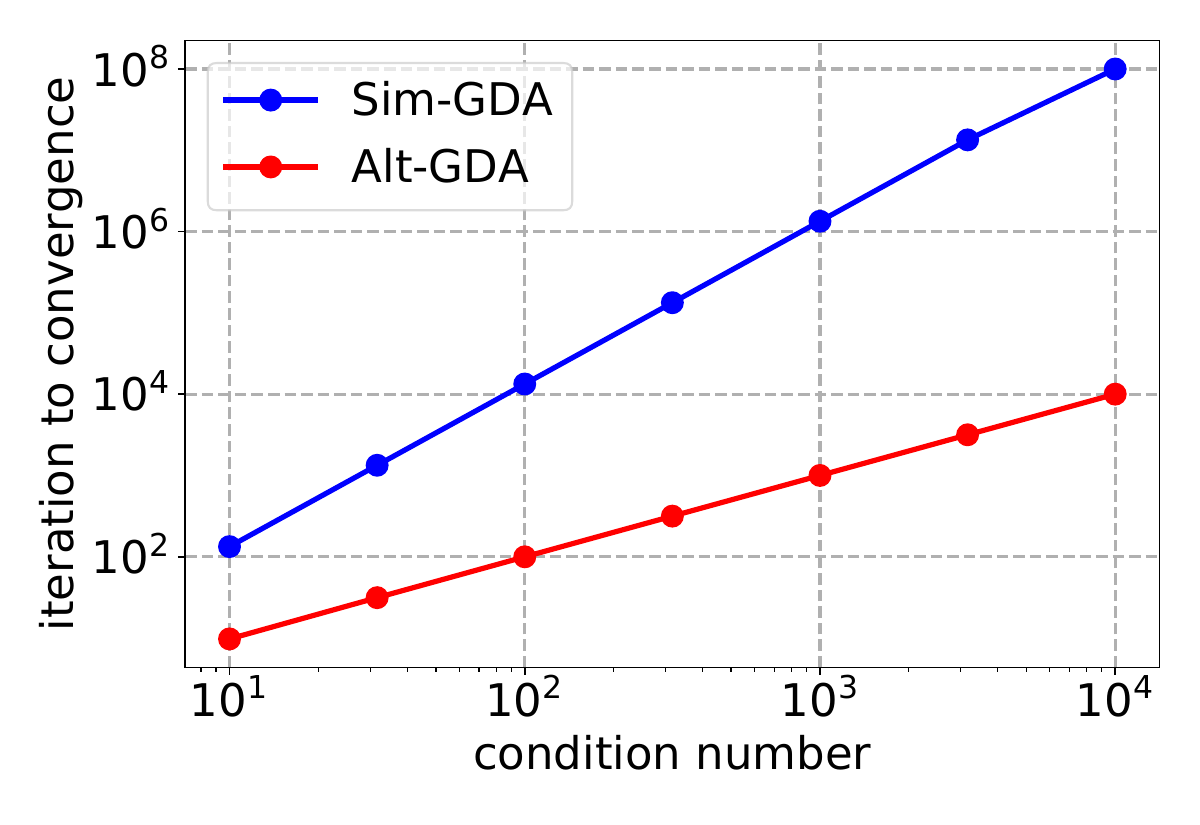}
	\vspace{-0.3cm}
	\caption{Certified iteration complexities of Sim-GDA and Alt-GDA for blinearly-coupled minimax games. \textbf{Observations:} (1) Sim-GDA converges with an iteration complexity of $\bigo(\kappa^2)$; (1) Alt-GDA achieves an improved rate of $\bigo(\kappa)$, which matches the local rate.}
	\label{fig:iqc}
	\vspace{-0.1cm}
\end{figure}

In Figure~\ref{fig:iqc}, we plot the IQC-certified iteration complexity as a function of condition number. We observe that the bound for Alt-GDA does improve upon that of Sim-GDA, especially when the condition number is large. In particular, the complexity of Alt-GDA scales
linearly with the condition number, suggesting its iteration complexity is $\bigo(\kappa)$. This implies that Alt-GDA does accelerate the convergence globally for this class of problem.

\section{RELATED WORK}
\begin{figure*}[t]
\centering
	\begin{subfigure}[b]{0.32\textwidth}
        \centering
        \includegraphics[width=\textwidth]{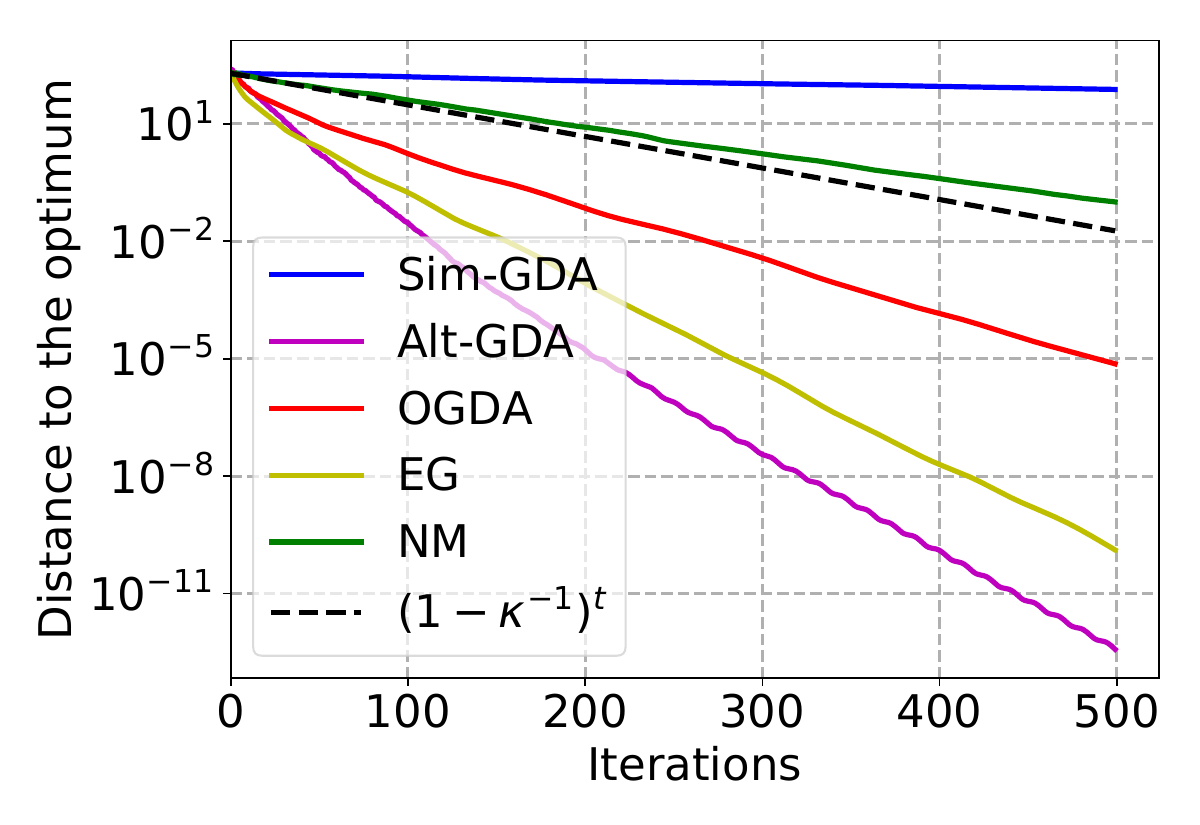}
	\end{subfigure}
	\begin{subfigure}[b]{0.32\textwidth}
        \centering
        \includegraphics[width=\textwidth]{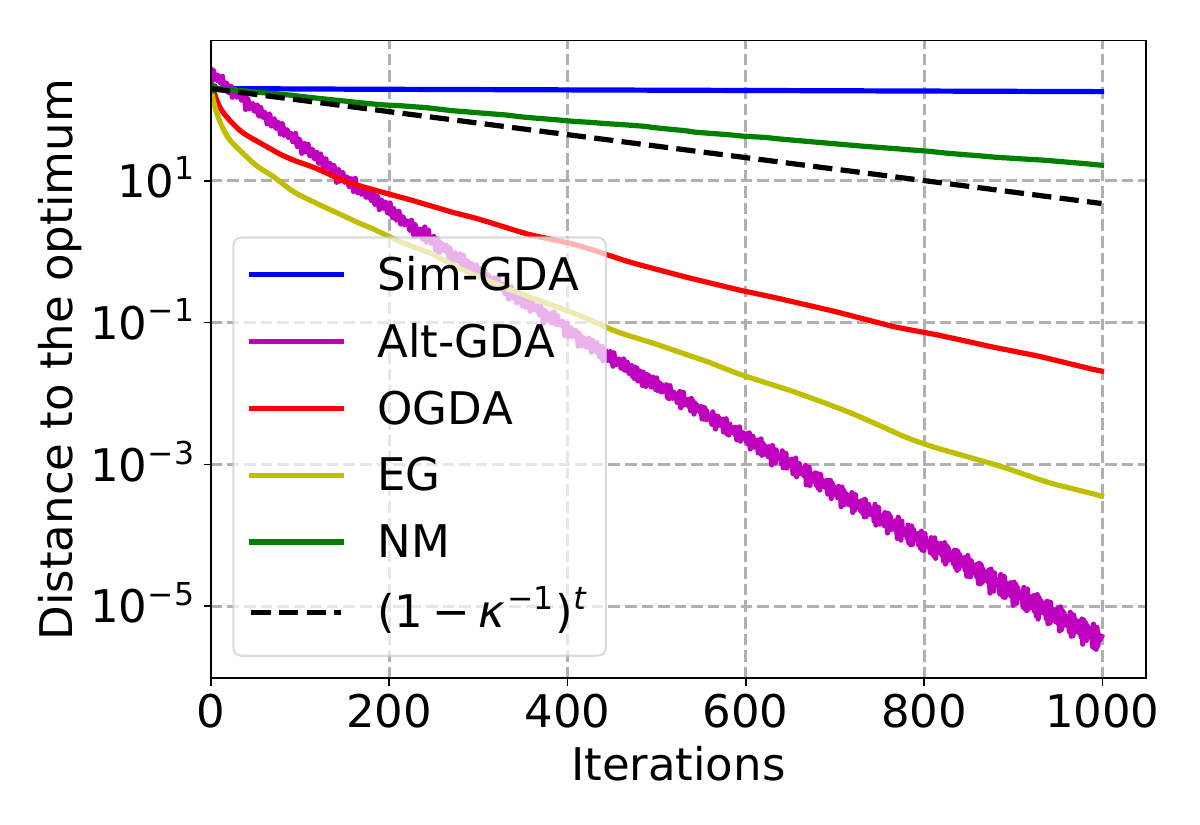}
	\end{subfigure}
	\begin{subfigure}[b]{0.32\textwidth}
        \centering
        \includegraphics[width=\textwidth]{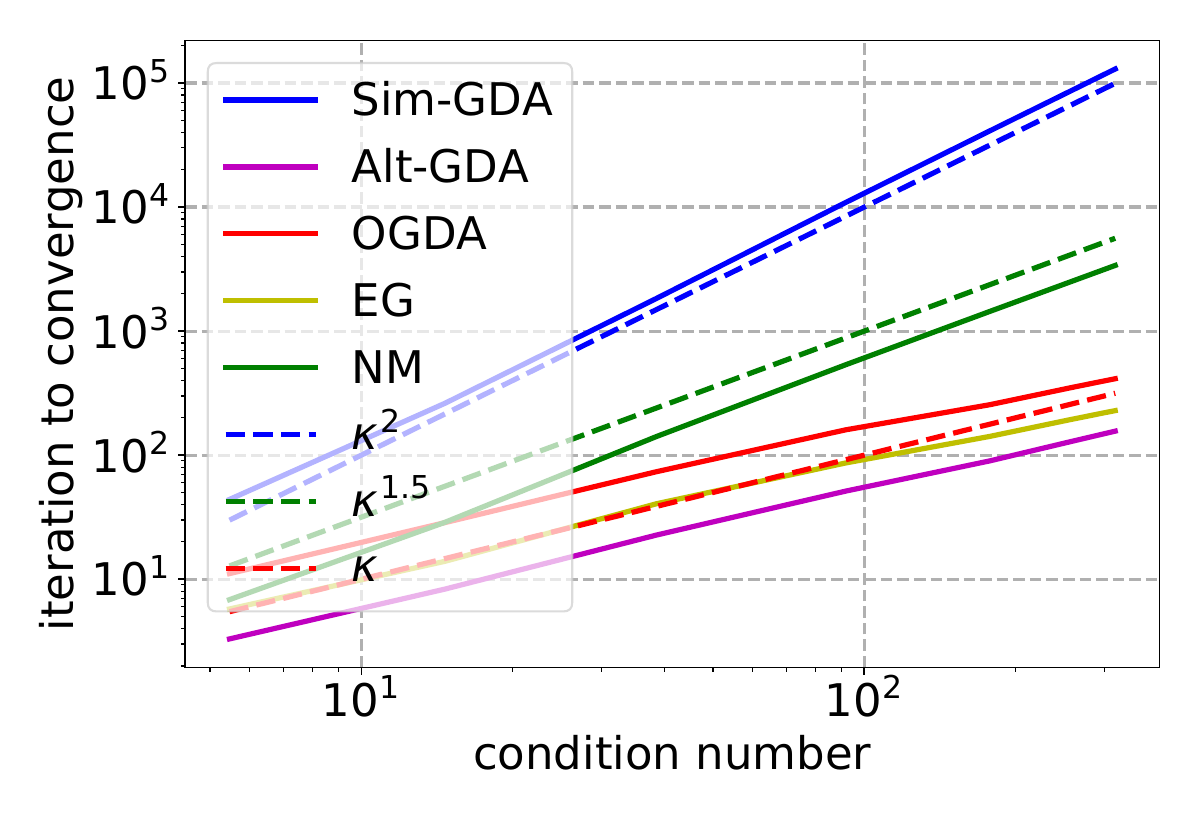}
	\end{subfigure}
	\vspace{-0.2cm}
	\caption{\textbf{Left:} Distance to the optimum as a function of training iterations. Alt-GDA accelerates Sim-GDA significantly on this particular quadratic minimax problem. In particular, its convergence rate is better than the worst-case rate of $1 - \kappa^{-1}$ with $\kappa = {\max\{|\lambda_i|\}}/{\min\{\realp(\lambda_i)\}}$ ($\kappa \approx 50$). Moreover, Alt-GDA outperforms OGDA and EG by a visible margin albeit the fact that they all have the $\bigo(\kappa)$ complexity bound. 
	\textbf{Middle:} A hard problem with entries of $\bB$ sampled from $\mathcal{N}(0, 1)$ ($\kappa \approx 300$).
	\textbf{Right:} Iterations to convergence for the all $5$ algorithms with tuned hyperparameters. The iteration complexity of Alt-GDA scales linearly with the condition number while that of Sim-GDA scales quadratically.}
	\label{fig:simulation}
	\vspace{-0.2cm}
\end{figure*}
The discussion of simultaneous and alternating updates in iterative algorithms dates back to the Jacobi and Gauss–Seidel methods in numerical linear algebra~\citep{saad2003iterative}. The Jacobi method makes simultaneous updates and is therefore naturally amenable to parallelization. On the other hand, the Gauss-Seidel method updates sequentially, so that each update leverages fresh information, and therefore is typically more stable and converges in fewer iterations. In minimax optimization, there is an analogous trade-off between simultaneous and alternating updates.

The discussion of alternating algorithms is lacking and is largely limited to simple bilinear games.
\cite{gidel2019negative} showed that Alt-GDA stays bounded and negative momentum with alternating updates converges linearly in bilinear games. Later, \cite{bailey2020finite} extended the analysis of Alt-GDA to no-regret online learning, albeit just for simple bilinear games.
\cite{zhang2019convergence} provided some evidence that alternating versions of many popular algorithms outperform their simultaneous counterpart in bilinear games.
Very recently, \cite{yang2020global} established the global convergence of Alt-GDA in a subclass of nonconvex-nonconcave objectives satisfying a so-called two-sided Polyak-\L{}ojasiewicz inequality. \cite{xu2020unified, boct2020alternating} proved convergence rates for alternating (proximal) GDA for nonconvex-concave minimax problems.
However, it remains unclear whether these alternating methods improve the convergence compared to their simultaneous counterparts in the above two settings.

By contrast, there is a large body of work on simultaneous methods in minimax optimization. For the strongly convex-strongly concave case, \cite{tseng1995linear} and \cite{nesterov2006solving} proved that their algorithms find an $\epsilon$-saddle point with a gradient complexity of $\bigo(\kappa\ln(1/\epsilon))$ using a variational inequality approach. Using a different approach, \cite{gidel2018variational} and \cite{mokhtari2020unified} derived the same convergence results for OGDA. Particularly, \cite{mokhtari2020unified} gave a unified analysis of OGDA and EG from the perspective of proximal point methods. Later, \cite{zhang2020unified} provided a unified and automated framework for analyzing various first-order methods using the theory of integral quadratic constraints from control theory. 
Very recently, \cite{ibrahim2019linear, zhang2019lower} established fine-grained lower complexity bounds among all the first-order algorithms in this setting, and these bounds were later achieved by~\cite{lin2020near, wang2020improved}. 
For the convex-concave setting, it is known that the optimal rate of convergence for first-order methods is $\bigo(1/T)$, and this rate is achieved by both the EG and OGDA algorithms~\citep{nemirovski2004prox, tseng2008accelerated, hsieh2019convergence, mokhtari2020convergence} for the averaged (ergodic) iterates. 
Later, \citep{golowich2020last, golowich2020tight} derived a $\bigo(1/\sqrt{T})$ bound for the last iterate of EG and OGDA.

\begin{figure*}[t]
\centering
    \begin{subfigure}[b]{0.195\textwidth}
        \centering
        \includegraphics[width=\textwidth]{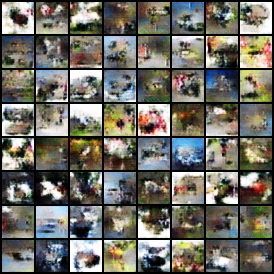}
        \vspace{-0.3cm}
        \caption{Sim-GDA}
        \label{subfig:sim-gda}
    \end{subfigure}
    \begin{subfigure}[b]{0.195\textwidth}
        \centering
        \includegraphics[width=\textwidth]{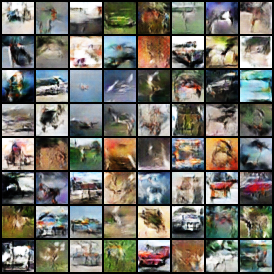}
        \vspace{-0.3cm}
        \caption{Alt-GDA}
        \label{subfig:alt-gda}
    \end{subfigure}
    \begin{subfigure}[b]{0.27\textwidth}
        \centering
        \includegraphics[width=\textwidth]{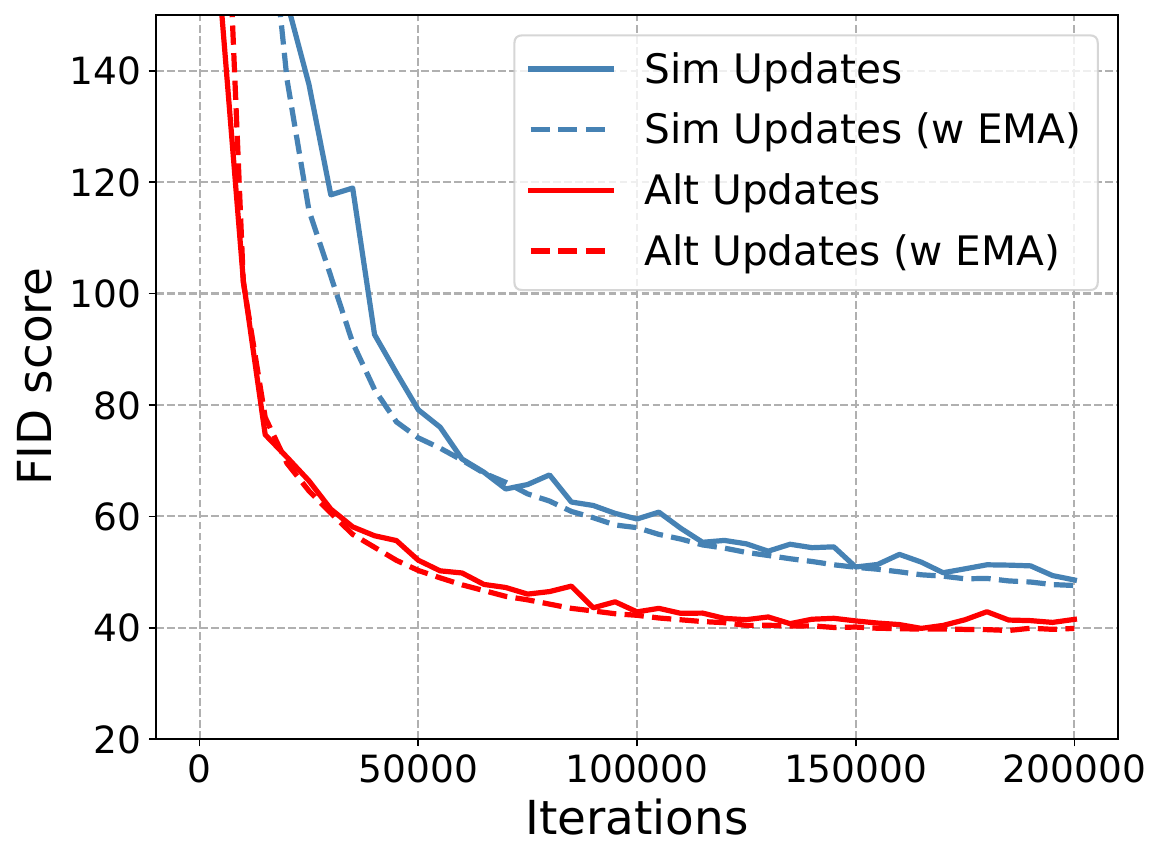}
        \vspace{-0.3cm}
        \caption{Trained with SGD}
        \label{subfig:sgd}
    \end{subfigure}
    \begin{subfigure}[b]{0.27\textwidth}
        \centering
        \includegraphics[width=\textwidth]{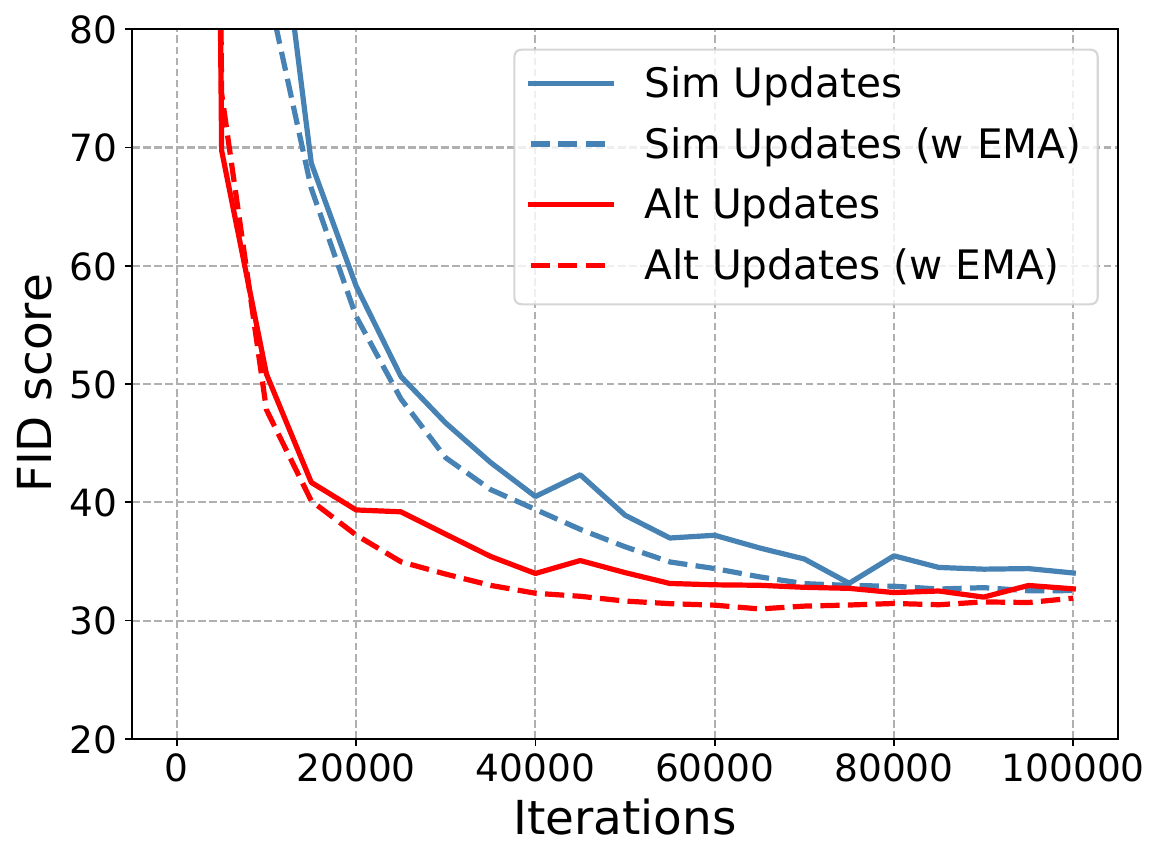}
        \vspace{-0.3cm}
        \caption{Trained with AMSGrad}
        \label{subfig:amsgrad}
    \end{subfigure}
    \vspace{-0.1cm}
	\caption{Comparisons between simultaneous and alternating algorithms on GAN training. \textbf{(a)} We train the DCGAN model on CIFAR-10 with simultaneous SGD . The samples are saved at iteration 30000. \textbf{(b)} We train the DCGAN model on CIFAR10 with alternating SGD. Again, the samples are saved at iteration 30000. \textbf{(c)} The curves of FID scores on CIFAR-10 with SGD. The dash lines are computed at exponential moving averaged models. \textbf{(d)} The curves of FID scores on CIFAR-10 with AMSGrad. In all settings, alternating version of the algorithms converges much faster than their simultaneous counterparts and achieve better FID scores in the end. } %
	\label{fig:gan}
	\vspace{-0.2cm}
\end{figure*}

\section{EXPERIMENTS}
\subsection{Quadratic Minimax Games}
In this section, we compare the performance of Alt-GDA with Sim-GDA along with other three popular algorithms (EG, OGDA and NM) so as to verify our theoretical results on the convergence rate of Alt-GDA. In particular, we focus on the following quadratic minimax problem:
\vspace{-0.2cm}
\begin{equation}
    \min_{\bx \in\real^d} \max_{\by \in \real^d} f(\bx, \by) = \frac{1}{2}\bx^\top \bA \bx + \bx^\top \bB \by - \frac{1}{2} \by^\top \bC \by
    \vspace{-0.1cm}
\end{equation}
where we set the dimension $d = 100$. We note both linear regression~\citep{du2019linear} and robust least squares~\citep{yang2020global} problems admit this minimax formulation.
The matrices $\bA$ and $\bC$ are set to have eigenvalues $\{\tfrac{1}{i}\}_{i=1}^{d}$. For matrix $\bB$, we set it to be a random matrix with entries sampling from a Gaussian distribution (either $\mathcal{N}(0, 0.01)$ or  $\mathcal{N}(0, 1)$). In the case of $\bB$ sampled from $\mathcal{N}(0, 1)$, the resulting gradient vector field has a strong rotational force since the off-diagonal blocks of its Jacobian dominates (see \eqref{eq:jacobian} in the Appendix).
For all algorithms, the iterates start with $\bx_0 = \mathbf{1}$ and $\by_0 = \mathbf{1}$. Figure~\ref{fig:simulation} shows that the distance to the optimum of Sim-GDA, Alt-GDA, OGDA, EG and NM\footnote{We implemented the simultaneous version of negative momentum (NM). For alternating NM, the optimal damping value of NM is roughly zero, making it the same algorithm as Alt-GDA.} versus the number of iterations for this problem. For all methods, we tune their hyperparameters by grid-search. We notice that all methods converge linearly to the optimum. As expected, Alt-GDA performs significantly better than Sim-GDA and yields a convergence rate that is better than its worst-case rate (black dashed line). Moreover, we find that Alt-GDA outperforms OGDA and EG by a visible margin. This is surprising, in that OGDA and EG take another memory buffer for accelerating the convergence.

Furthermore, we study how the convergence rates (or iteration complexities) scale with the condition numbers. To this end, we randomly sample\footnote{We let the eigenvalues of matrices $\bA$ and $\bC$ be $\{ \tfrac{1}{n_i}\}_{i=1}^d$ where $n_i$ are evenly spaced from $1$ to $N$, where $N$ is in $[\sqrt{10}, 10^3]$. We sample all entries of $\bB$ from standard normal distribution $\mathcal{N}(0, 1)$ and then normalize it. } matrices $\mathbf{A}, \mathbf{B}, \mathbf{C}$ and compute the condition number by $\kappa = \frac{\max\{|\lambda_i|\}}{\min\{\realp(\lambda_i)\}}$ where $\lambda_i$ are eigenvalues of the Jacobian $\jacobian$ of the gradient vector field in \eqref{eq:jacobian}. Once we have all these three matrices, we can compute the spectral radius $\rho$ of all algorithms with tuned step-sizes and momentum value. We plot $-1/\log(\rho)$ versus the condition number $\kappa$ in Figure~\ref{fig:simulation} (right) to get a sense of how the relative iteration complexity scales as a function of condition number. We find that the iteration complexity of Alt-GDA scales linearly with the condition number, matching our prediction in Corollary~\ref{cor:alt-gda}. On the other hand, Sim-GDA takes roughly $\kappa^2$ iterations to convergence, as predicted in Theorem~\ref{thm:sim-gda}.
In addition, Alt-GDA is slightly better than OGDA and EG as its curve is below that of OGDA and EG, albeit with the same slope.

\subsection{Generative Adversarial Networks}
In this section, we investigate the effect of alternating updates on training generative adversarial networks. The purpose of this section is to show that the insights gained from our analyses carry over to GAN training despite the fact that the GAN objective is generally nonconvex-nonconcave. In addition, we note that while GAN training is a stochastic problem, stochastic problems are sometimes in a curvature-dominated regime where the convergence behavior resembles that of the deterministic problems~\citep{zhang2019algorithmic}.

We first compare alternating algorithms with their simultaneous counterparts on CIFAR10~\citep{krizhevsky2009learning} image generation task with the WGAN-GP~\citep{gulrajani2017improved} objective and a DCGAN~\citep{radford2015unsupervised} architecture. In particular, we choose SGD and AMSGrad~\citep{reddi2019convergence} as our base optimizers. For more implementation details, please see Appendix~\ref{app:imp}.
We evaluate all algorithms with Fréchet Inception Distance\footnote{Inception score~\citep{salimans2016improved} is also a popular metric, however it was shown by~\cite{chavdarova2020taming} that it is less consistent with the sample quality, so we instead use FID score here.} (FID)~\citep{heusel2017gans}. 
Figure~\ref{subfig:sgd} and~\ref{subfig:amsgrad} summarize our results. With SGD as our optimizer\footnote{We also include the results with exponential moving average (EMA).}, we observe that alternating SGD not only converges faster, but also converges to a better point with lower FID score. Although both alternating version and simultaneous version of AMSGrad converges to models with similar FID scores, the alternating version again converges with many fewer iterations, matching our prediction. In addition, we generate samples from trained Generators at iteration 30000 with SGD optimizer (see Figure~\ref{subfig:sim-gda} and~\ref{subfig:alt-gda}). It is easy to see that the model trained with alternating updates generates better samples given the same compute budget.

\begin{figure}[t]
\centering
    \begin{subfigure}[b]{0.46\columnwidth}
        \centering
        \includegraphics[width=\columnwidth]{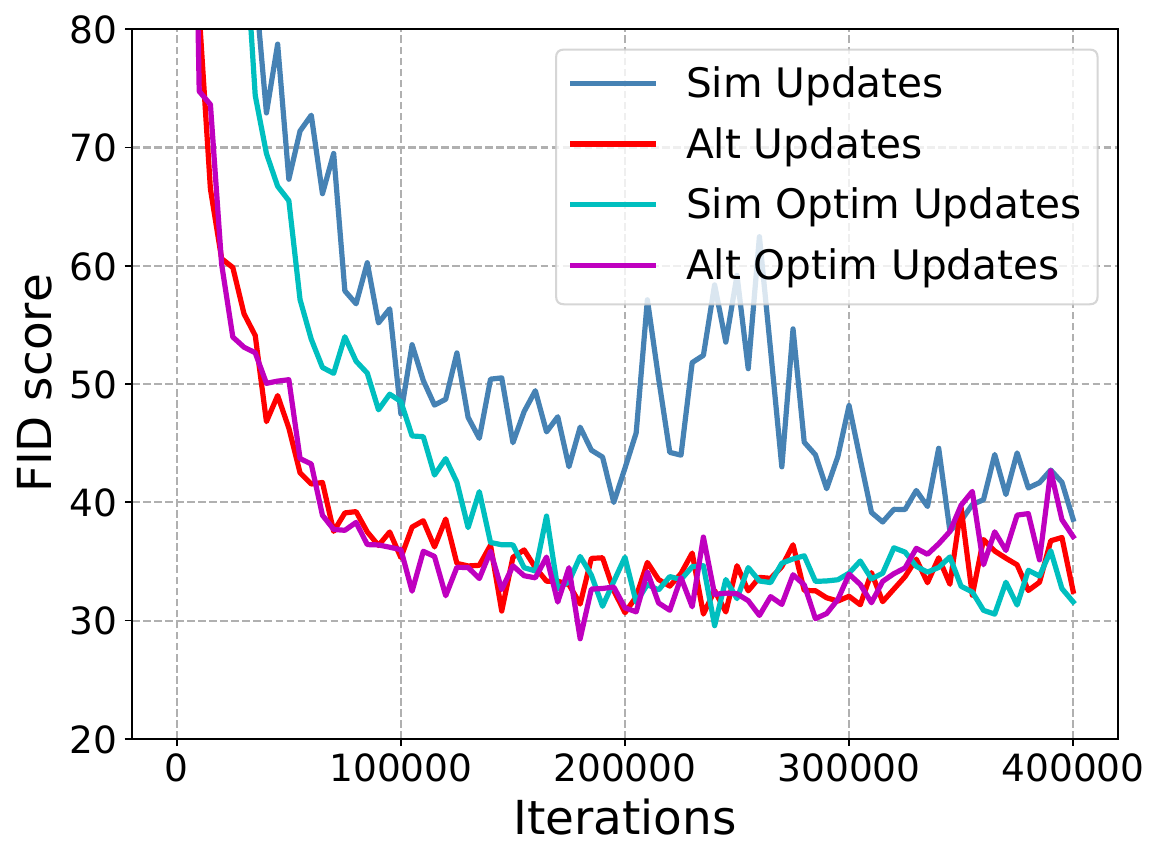}
        \vspace{-0.6cm}
        \caption{SGD}
	\end{subfigure}
	\begin{subfigure}[b]{0.46\columnwidth}
        \centering
        \includegraphics[width=\columnwidth]{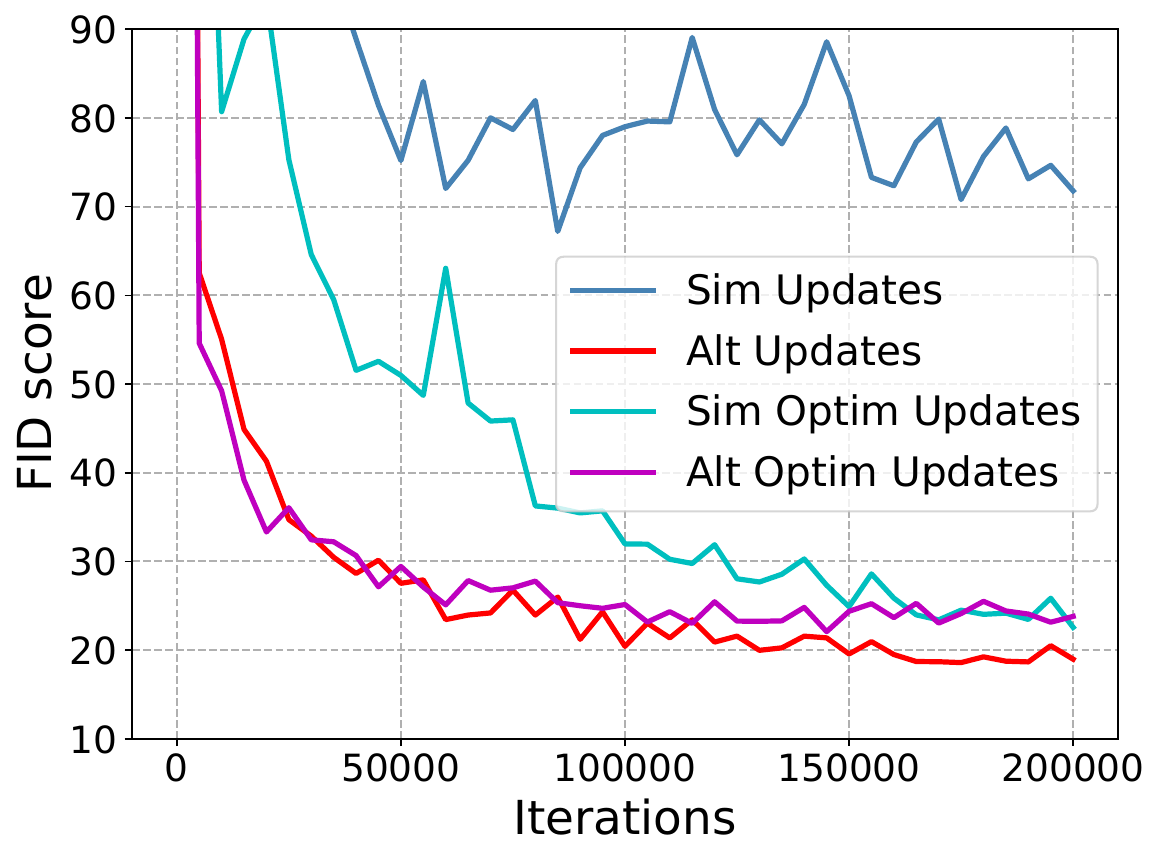}
        \vspace{-0.6cm}
        \caption{AMSGrad}
	\end{subfigure}
	\vspace{-0.2cm}
	\caption{\textbf{(a)} ResNet model trained with SGD on CIFAR-10. \textbf{(b)} ResNet model trained with AMSGrad on CIFAR-10. Alternating algorithms dominate simultaneous ones. More interestingly, the use of optimism does \emph{not} help for alternating algorithms.}
	\label{fig:resnet-gan}
	\vspace{-0.3cm}
\end{figure}
Furthermore, we compare simultaneous methods and alternating methods on a deep ResNet~\citep{miyato2018spectral}. We also include optimistic updates~\citep{daskalakis2018training} in the training, which is the key component of OGDA\footnote{See Appendix~\ref{app:imp} for detailed update rule.}. We report all results in Figure~\ref{fig:resnet-gan}. The first observation is that alternating algorithms take fewer iterations to converge regardless of whether optimism is used, and sometimes converge to models with better FID scores (similar to DCGAN results). Second, we observe that the use of optimism only helps for simultaneous algorithms, suggesting that alternating updates and optimistic updates play similar roles in improving GAN training. This could be explained by our theoretical results that Alt-GDA enjoys a similar convergence rate to OGDA.

\section{CONCLUSION}
In this paper, we take an important step towards understanding alternating algorithms in minimax optimization by analyzing Alt-GDA in three distinct settings. In particular, we show theoretically that Alt-GDA outperforms its simultaneous counterpart by a big margin in all three settings. Unexpectedly, Alt-GDA achieves a near-optimal convergence rate locally for strongly convex-strongly concave smooth minimax games, matching the known coarse lower bound. 
Moreover, the acceleration effect of Alt-GDA remains when the minimax problem has only strong concavity in the dual variables.

Our numerical simulations on toy quadratic games verified our claims. Further, we demonstrate empirically that alternating updates could significantly speed up GAN training though GAN objective is generally nonconvex-nonconcave. More interestingly, we show that the use of optimism only helps for simultaneous algorithms. We believe that the default use of alternating update rule in GAN training was an important reason for its success.

\bibliography{reference}

\begin{thebibliography}{67}
\providecommand{\natexlab}[1]{#1}
\providecommand{\url}[1]{\texttt{#1}}
\expandafter\ifx\csname urlstyle\endcsname\relax
  \providecommand{\doi}[1]{doi: #1}\else
  \providecommand{\doi}{doi: \begingroup \urlstyle{rm}\Url}\fi

\bibitem[Arjovsky et~al.(2017)Arjovsky, Chintala, and
  Bottou]{arjovsky2017wasserstein}
Martin Arjovsky, Soumith Chintala, and L{\'e}on Bottou.
\newblock Wasserstein generative adversarial networks.
\newblock In \emph{International Conference on Machine Learning}, pages
  214--223, 2017.

\bibitem[Azizian et~al.(2020{\natexlab{a}})Azizian, Mitliagkas, Lacoste-Julien,
  and Gidel]{azizian2020tight}
Wa{\"\i}ss Azizian, Ioannis Mitliagkas, Simon Lacoste-Julien, and Gauthier
  Gidel.
\newblock A tight and unified analysis of gradient-based methods for a whole
  spectrum of differentiable games.
\newblock In \emph{International Conference on Artificial Intelligence and
  Statistics}, pages 2863--2873, 2020{\natexlab{a}}.

\bibitem[Azizian et~al.(2020{\natexlab{b}})Azizian, Scieur, Mitliagkas,
  Lacoste-Julien, and Gidel]{azizian2020accelerating}
Waïss Azizian, Damien Scieur, Ioannis Mitliagkas, Simon Lacoste-Julien, and
  Gauthier Gidel.
\newblock Accelerating smooth games by manipulating spectral shapes.
\newblock In \emph{Proceedings of the Twenty Third International Conference on
  Artificial Intelligence and Statistics}, pages 1705--1715,
  2020{\natexlab{b}}.

\bibitem[Bailey and Piliouras(2018)]{bailey2018multiplicative}
James~P Bailey and Georgios Piliouras.
\newblock Multiplicative weights update in zero-sum games.
\newblock In \emph{Proceedings of the 2018 ACM Conference on Economics and
  Computation}, pages 321--338, 2018.

\bibitem[Bailey et~al.(2020)Bailey, Gidel, and Piliouras]{bailey2020finite}
James~P Bailey, Gauthier Gidel, and Georgios Piliouras.
\newblock Finite regret and cycles with fixed step-size via alternating
  gradient descent-ascent.
\newblock In \emph{Conference on Learning Theory}, pages 391--407. PMLR, 2020.

\bibitem[Ba{\c{s}}ar and Olsder(1998)]{bacsar1998dynamic}
Tamer Ba{\c{s}}ar and Geert~Jan Olsder.
\newblock \emph{Dynamic noncooperative game theory}.
\newblock SIAM, 1998.

\bibitem[Ben-Tal et~al.(2009)Ben-Tal, El~Ghaoui, and Nemirovski]{ben2009robust}
Aharon Ben-Tal, Laurent El~Ghaoui, and Arkadi Nemirovski.
\newblock \emph{Robust optimization}, volume~28.
\newblock Princeton University Press, 2009.

\bibitem[Bertsekas(1997)]{bertsekas1997nonlinear}
Dimitri~P Bertsekas.
\newblock Nonlinear programming.
\newblock \emph{Journal of the Operational Research Society}, 48\penalty0
  (3):\penalty0 334--334, 1997.

\bibitem[Bo{\c{t}} and B{\"o}hm(2020)]{boct2020alternating}
Radu~Ioan Bo{\c{t}} and Axel B{\"o}hm.
\newblock Alternating proximal-gradient steps for (stochastic)
  nonconvex-concave minimax problems.
\newblock \emph{arXiv preprint arXiv:2007.13605}, 2020.

\bibitem[Chambolle and Pock(2011)]{chambolle2011first}
Antonin Chambolle and Thomas Pock.
\newblock A first-order primal-dual algorithm for convex problems with
  applications to imaging.
\newblock \emph{Journal of mathematical imaging and vision}, 40\penalty0
  (1):\penalty0 120--145, 2011.

\bibitem[Chavdarova et~al.(2021)Chavdarova, Pagliardini, Stich, Fleuret, and
  Jaggi]{chavdarova2020taming}
Tatjana Chavdarova, Matteo Pagliardini, Sebastian~U Stich, Fran{\c{c}}ois
  Fleuret, and Martin Jaggi.
\newblock Taming {GAN}s with lookahead-minmax.
\newblock In \emph{International Conference on Learning Representations}, 2021.

\bibitem[Daskalakis and Panageas(2018)]{daskalakis2018last}
Constantinos Daskalakis and Ioannis Panageas.
\newblock Last-iterate convergence: Zero-sum games and constrained min-max
  optimization.
\newblock In \emph{10th Innovations in Theoretical Computer Science Conference
  (ITCS 2019)}. Schloss Dagstuhl-Leibniz-Zentrum fuer Informatik, 2018.

\bibitem[Daskalakis et~al.(2018)Daskalakis, Ilyas, Syrgkanis, and
  Zeng]{daskalakis2018training}
Constantinos Daskalakis, Andrew Ilyas, Vasilis Syrgkanis, and Haoyang Zeng.
\newblock Training gans with optimism.
\newblock In \emph{International Conference on Learning Representations}, 2018.

\bibitem[Du and Hu(2019)]{du2019linear}
Simon~S Du and Wei Hu.
\newblock Linear convergence of the primal-dual gradient method for
  convex-concave saddle point problems without strong convexity.
\newblock In \emph{The 22nd International Conference on Artificial Intelligence
  and Statistics}, pages 196--205, 2019.

\bibitem[Du et~al.(2017)Du, Chen, Li, Xiao, and Zhou]{du2017stochastic}
Simon~S Du, Jianshu Chen, Lihong Li, Lin Xiao, and Dengyong Zhou.
\newblock Stochastic variance reduction methods for policy evaluation.
\newblock In \emph{International Conference on Machine Learning}, pages
  1049--1058, 2017.

\bibitem[Fiez and Ratliff(2021)]{fiez2020gradient}
Tanner Fiez and Lillian~J Ratliff.
\newblock Local convergence analysis of gradient descent ascent with finite
  timescale separation.
\newblock In \emph{International Conference on Learning Representations}, 2021.

\bibitem[Gidel et~al.(2019{\natexlab{a}})Gidel, Berard, Vignoud, Vincent, and
  Lacoste-Julien]{gidel2018variational}
Gauthier Gidel, Hugo Berard, Gaëtan Vignoud, Pascal Vincent, and Simon
  Lacoste-Julien.
\newblock A variational inequality perspective on generative adversarial
  networks.
\newblock 2019{\natexlab{a}}.

\bibitem[Gidel et~al.(2019{\natexlab{b}})Gidel, Hemmat, Pezeshki, Le~Priol,
  Huang, Lacoste-Julien, and Mitliagkas]{gidel2019negative}
Gauthier Gidel, Reyhane~Askari Hemmat, Mohammad Pezeshki, R{\'e}mi Le~Priol,
  Gabriel Huang, Simon Lacoste-Julien, and Ioannis Mitliagkas.
\newblock Negative momentum for improved game dynamics.
\newblock In \emph{The 22nd International Conference on Artificial Intelligence
  and Statistics}, pages 1802--1811, 2019{\natexlab{b}}.

\bibitem[Golowich et~al.(2020{\natexlab{a}})Golowich, Pattathil, and
  Daskalakis]{golowich2020tight}
Noah Golowich, Sarath Pattathil, and Constantinos Daskalakis.
\newblock Tight last-iterate convergence rates for no-regret learning in
  multi-player games.
\newblock \emph{Advances in Neural Information Processing Systems}, 33,
  2020{\natexlab{a}}.

\bibitem[Golowich et~al.(2020{\natexlab{b}})Golowich, Pattathil, Daskalakis,
  and Ozdaglar]{golowich2020last}
Noah Golowich, Sarath Pattathil, Constantinos Daskalakis, and Asuman Ozdaglar.
\newblock Last iterate is slower than averaged iterate in smooth convex-concave
  saddle point problems.
\newblock In \emph{Proceedings of Thirty Third Conference on Learning Theory},
  pages 1758--1784, 2020{\natexlab{b}}.

\bibitem[Goodfellow et~al.(2014)Goodfellow, Pouget-Abadie, Mirza, Xu,
  Warde-Farley, Ozair, Courville, and Bengio]{goodfellow2014generative}
Ian Goodfellow, Jean Pouget-Abadie, Mehdi Mirza, Bing Xu, David Warde-Farley,
  Sherjil Ozair, Aaron Courville, and Yoshua Bengio.
\newblock Generative adversarial nets.
\newblock In \emph{Advances in neural information processing systems}, pages
  2672--2680, 2014.

\bibitem[Gulrajani et~al.(2017)Gulrajani, Ahmed, Arjovsky, Dumoulin, and
  Courville]{gulrajani2017improved}
Ishaan Gulrajani, Faruk Ahmed, Martin Arjovsky, Vincent Dumoulin, and Aaron~C
  Courville.
\newblock Improved training of wasserstein gans.
\newblock In \emph{Advances in neural information processing systems}, pages
  5767--5777, 2017.

\bibitem[Harker and Pang(1990)]{harker1990finite}
Patrick~T Harker and Jong-Shi Pang.
\newblock Finite-dimensional variational inequality and nonlinear
  complementarity problems: a survey of theory, algorithms and applications.
\newblock \emph{Mathematical programming}, 48\penalty0 (1-3):\penalty0
  161--220, 1990.

\bibitem[Heusel et~al.(2017)Heusel, Ramsauer, Unterthiner, Nessler, and
  Hochreiter]{heusel2017gans}
Martin Heusel, Hubert Ramsauer, Thomas Unterthiner, Bernhard Nessler, and Sepp
  Hochreiter.
\newblock Gans trained by a two time-scale update rule converge to a local nash
  equilibrium.
\newblock In \emph{Advances in neural information processing systems}, pages
  6626--6637, 2017.

\bibitem[Hsieh et~al.(2019)Hsieh, Iutzeler, Malick, and
  Mertikopoulos]{hsieh2019convergence}
Yu-Guan Hsieh, Franck Iutzeler, J{\'e}r{\^o}me Malick, and Panayotis
  Mertikopoulos.
\newblock On the convergence of single-call stochastic extra-gradient methods.
\newblock In \emph{Advances in Neural Information Processing Systems}, pages
  6938--6948, 2019.

\bibitem[Ibrahim et~al.(2020)Ibrahim, Azizian, Gidel, and
  Mitliagkas]{ibrahim2019linear}
Adam Ibrahim, Wa{\i}ss Azizian, Gauthier Gidel, and Ioannis Mitliagkas.
\newblock Linear lower bounds and conditioning of differentiable games.
\newblock In \emph{International Conference on Machine Learning}, pages
  4583--4593. PMLR, 2020.

\bibitem[Korpelevich(1976)]{korpelevich1976extragradient}
Galina~M Korpelevich.
\newblock The extragradient method for finding saddle points and other
  problems.
\newblock \emph{Matecon}, 12:\penalty0 747--756, 1976.

\bibitem[Krizhevsky et~al.(2009)]{krizhevsky2009learning}
Alex Krizhevsky et~al.
\newblock Learning multiple layers of features from tiny images.
\newblock 2009.

\bibitem[Lee et~al.(2016)Lee, Simchowitz, Jordan, and Recht]{lee2016gradient}
Jason~D Lee, Max Simchowitz, Michael~I Jordan, and Benjamin Recht.
\newblock Gradient descent only converges to minimizers.
\newblock In \emph{Conference on learning theory}, pages 1246--1257, 2016.

\bibitem[Lee et~al.(2017)Lee, Panageas, Piliouras, Simchowitz, Jordan, and
  Recht]{lee2017first}
Jason~D Lee, Ioannis Panageas, Georgios Piliouras, Max Simchowitz, Michael~I
  Jordan, and Benjamin Recht.
\newblock First-order methods almost always avoid saddle points.
\newblock \emph{arXiv preprint arXiv:1710.07406}, 2017.

\bibitem[Lessard et~al.(2016)Lessard, Recht, and Packard]{lessard2016analysis}
Laurent Lessard, Benjamin Recht, and Andrew Packard.
\newblock Analysis and design of optimization algorithms via integral quadratic
  constraints.
\newblock \emph{SIAM Journal on Optimization}, 26\penalty0 (1):\penalty0
  57--95, 2016.

\bibitem[Letcher et~al.(2019)Letcher, Balduzzi, Racaniere, Martens, Foerster,
  Tuyls, and Graepel]{letcher2019differentiable}
Alistair Letcher, David Balduzzi, S{\'e}bastien Racaniere, James Martens,
  Jakob~N Foerster, Karl Tuyls, and Thore Graepel.
\newblock Differentiable game mechanics.
\newblock \emph{J. Mach. Learn. Res.}, 20:\penalty0 84--1, 2019.

\bibitem[Liang and Stokes(2019)]{liang2019interaction}
Tengyuan Liang and James Stokes.
\newblock Interaction matters: A note on non-asymptotic local convergence of
  generative adversarial networks.
\newblock In \emph{The 22nd International Conference on Artificial Intelligence
  and Statistics}, pages 907--915, 2019.

\bibitem[Lin et~al.(2020)Lin, Jin, and Jordan]{lin2020near}
Tianyi Lin, Chi Jin, and Michael~I. Jordan.
\newblock Near-optimal algorithms for minimax optimization.
\newblock In \emph{Proceedings of Thirty Third Conference on Learning Theory},
  pages 2738--2779, 2020.

\bibitem[Madry et~al.(2018)Madry, Makelov, Schmidt, Tsipras, and
  Vladu]{madry2018towards}
Aleksander Madry, Aleksandar Makelov, Ludwig Schmidt, Dimitris Tsipras, and
  Adrian Vladu.
\newblock Towards deep learning models resistant to adversarial attacks.
\newblock In \emph{International Conference on Learning Representations}, 2018.

\bibitem[Mertikopoulos et~al.(2019)Mertikopoulos, Lecouat, Zenati, Foo,
  Chandrasekhar, and Piliouras]{mertikopoulos2018optimistic}
Panayotis Mertikopoulos, Bruno Lecouat, Houssam Zenati, Chuan-Sheng Foo, Vijay
  Chandrasekhar, and Georgios Piliouras.
\newblock Optimistic mirror descent in saddle-point problems: Going the
  extra(-gradient) mile.
\newblock In \emph{International Conference on Learning Representations}, 2019.

\bibitem[Mescheder et~al.(2017)Mescheder, Nowozin, and
  Geiger]{mescheder2017numerics}
Lars Mescheder, Sebastian Nowozin, and Andreas Geiger.
\newblock The numerics of gans.
\newblock In \emph{Advances in Neural Information Processing Systems}, pages
  1825--1835, 2017.

\bibitem[Miyato et~al.(2018)Miyato, Kataoka, Koyama, and
  Yoshida]{miyato2018spectral}
Takeru Miyato, Toshiki Kataoka, Masanori Koyama, and Yuichi Yoshida.
\newblock Spectral normalization for generative adversarial networks.
\newblock In \emph{International Conference on Learning Representations}, 2018.

\bibitem[Mokhtari et~al.(2020{\natexlab{a}})Mokhtari, Ozdaglar, and
  Pattathil]{mokhtari2020unified}
Aryan Mokhtari, Asuman Ozdaglar, and Sarath Pattathil.
\newblock A unified analysis of extra-gradient and optimistic gradient methods
  for saddle point problems: Proximal point approach.
\newblock In \emph{International Conference on Artificial Intelligence and
  Statistics}, pages 1497--1507, 2020{\natexlab{a}}.

\bibitem[Mokhtari et~al.(2020{\natexlab{b}})Mokhtari, Ozdaglar, and
  Pattathil]{mokhtari2020convergence}
Aryan Mokhtari, Asuman~E Ozdaglar, and Sarath Pattathil.
\newblock Convergence rate of o(1/k) for optimistic gradient and extragradient
  methods in smooth convex-concave saddle point problems.
\newblock \emph{SIAM Journal on Optimization}, 30\penalty0 (4):\penalty0
  3230--3251, 2020{\natexlab{b}}.

\bibitem[Nemirovski(2004)]{nemirovski2004prox}
Arkadi Nemirovski.
\newblock Prox-method with rate of convergence o (1/t) for variational
  inequalities with lipschitz continuous monotone operators and smooth
  convex-concave saddle point problems.
\newblock \emph{SIAM Journal on Optimization}, 15\penalty0 (1):\penalty0
  229--251, 2004.

\bibitem[Nesterov and Scrimali(2011)]{nesterov2006solving}
Yurii Nesterov and Laura Scrimali.
\newblock Solving strongly monotone variational and quasi-variational
  inequalities.
\newblock \emph{Discrete \& Continuous Dynamical Systems-A}, 31\penalty0
  (4):\penalty0 1383, 2011.

\bibitem[Netzer et~al.(2011)Netzer, Wang, Coates, Bissacco, Wu, and
  Ng]{netzer2011reading}
Yuval Netzer, Tao Wang, Adam Coates, Alessandro Bissacco, Bo~Wu, and Andrew~Y
  Ng.
\newblock Reading digits in natural images with unsupervised feature learning.
\newblock 2011.

\bibitem[Popov(1980)]{popov1980modification}
Leonid~Denisovich Popov.
\newblock A modification of the arrow-hurwicz method for search of saddle
  points.
\newblock \emph{Mathematical notes of the Academy of Sciences of the USSR},
  28\penalty0 (5):\penalty0 845--848, 1980.

\bibitem[Radford et~al.(2015)Radford, Metz, and
  Chintala]{radford2015unsupervised}
Alec Radford, Luke Metz, and Soumith Chintala.
\newblock Unsupervised representation learning with deep convolutional
  generative adversarial networks.
\newblock \emph{arXiv preprint arXiv:1511.06434}, 2015.

\bibitem[Rakhlin and Sridharan(2013)]{rakhlin2013optimization}
Alexander Rakhlin and Karthik Sridharan.
\newblock Optimization, learning, and games with predictable sequences.
\newblock In \emph{Advances in Neural Information Processing Systems}, pages
  3066--3074, 2013.

\bibitem[Reddi et~al.(2018)Reddi, Kale, and Kumar]{reddi2019convergence}
Sashank~J. Reddi, Satyen Kale, and Sanjiv Kumar.
\newblock On the convergence of adam and beyond.
\newblock In \emph{International Conference on Learning Representations}, 2018.

\bibitem[Roughgarden(2010)]{roughgarden2010algorithmic}
Tim Roughgarden.
\newblock Algorithmic game theory.
\newblock \emph{Communications of the ACM}, 53\penalty0 (7):\penalty0 78--86,
  2010.

\bibitem[Saad(2003)]{saad2003iterative}
Yousef Saad.
\newblock \emph{Iterative methods for sparse linear systems}.
\newblock SIAM, 2003.

\bibitem[Salimans et~al.(2016)Salimans, Goodfellow, Zaremba, Cheung, Radford,
  and Chen]{salimans2016improved}
Tim Salimans, Ian~J Goodfellow, Wojciech Zaremba, Vicki Cheung, Alec Radford,
  and Xi~Chen.
\newblock Improved techniques for training gans.
\newblock In \emph{NIPS}, 2016.

\bibitem[Schaefer and Anandkumar(2019)]{florian2019competitive}
Florian Schaefer and Anima Anandkumar.
\newblock Competitive gradient descent.
\newblock In \emph{Advances in Neural Information Processing Systems}, 2019.

\bibitem[Tseng(1995)]{tseng1995linear}
Paul Tseng.
\newblock On linear convergence of iterative methods for the variational
  inequality problem.
\newblock \emph{Journal of Computational and Applied Mathematics}, 60\penalty0
  (1-2):\penalty0 237--252, 1995.

\bibitem[Tseng(2008)]{tseng2008accelerated}
Paul Tseng.
\newblock On accelerated proximal gradient methods for convex-concave
  optimization.
\newblock \emph{submitted to SIAM Journal on Optimization}, 2\penalty0 (3),
  2008.

\bibitem[von Neumann(1928)]{neumann1928theorie}
John von Neumann.
\newblock Zur theorie der gesellschaftsspiele.
\newblock \emph{Mathematische annalen}, 100\penalty0 (1):\penalty0 295--320,
  1928.

\bibitem[von Neumann and Morgenstern(1944)]{von1944theory}
John von Neumann and Oskar Morgenstern.
\newblock Theory of games and economic behavior.
\newblock 1944.

\bibitem[Wang and Li(2020)]{wang2020improved}
Yuanhao Wang and Jian Li.
\newblock Improved algorithms for convex-concave minimax optimization.
\newblock In \emph{Advances in Neural Information Processing Systems}, 2020.

\bibitem[Wang et~al.(2019)Wang, Zhang, and Ba]{wang2019solving}
Yuanhao Wang, Guodong Zhang, and Jimmy Ba.
\newblock On solving minimax optimization locally: A follow-the-ridge approach.
\newblock In \emph{International Conference on Learning Representations}, 2019.

\bibitem[Xie et~al.(2021)Xie, Han, and Zhang]{xie2021dippa}
Guangzeng Xie, Yuze Han, and Zhihua Zhang.
\newblock Dippa: An improved method for bilinear saddle point problems.
\newblock \emph{arXiv preprint arXiv:2103.08270}, 2021.

\bibitem[Xu et~al.(2020)Xu, Zhang, Xu, and Lan]{xu2020unified}
Zi~Xu, Huiling Zhang, Yang Xu, and Guanghui Lan.
\newblock A unified single-loop alternating gradient projection algorithm for
  nonconvex-concave and convex-nonconcave minimax problems.
\newblock \emph{arXiv preprint arXiv:2006.02032}, 2020.

\bibitem[Yang et~al.(2020{\natexlab{a}})Yang, Kiyavash, and He]{yang2020global}
Junchi Yang, Negar Kiyavash, and Niao He.
\newblock Global convergence and variance-reduced optimization for a class of
  nonconvex-nonconcave minimax problems.
\newblock In \emph{Advances in Neural Information Processing Systems},
  2020{\natexlab{a}}.

\bibitem[Yang et~al.(2020{\natexlab{b}})Yang, Zhang, Kiyavash, and
  He]{yang2020catalyst}
Junchi Yang, Siqi Zhang, Negar Kiyavash, and Niao He.
\newblock A catalyst framework for minimax optimization.
\newblock \emph{Advances in Neural Information Processing Systems}, 33,
  2020{\natexlab{b}}.

\bibitem[Yang et~al.(2020{\natexlab{c}})Yang, Nachum, Dai, Li, and
  Schuurmans]{yang2020off}
Mengjiao Yang, Ofir Nachum, Bo~Dai, Lihong Li, and Dale Schuurmans.
\newblock Off-policy evaluation via the regularized lagrangian.
\newblock \emph{Advances in Neural Information Processing Systems}, 33,
  2020{\natexlab{c}}.

\bibitem[Zhang and Wang(2021)]{zhang2020suboptimality}
Guodong Zhang and Yuanhao Wang.
\newblock On the suboptimality of negative momentum for minimax optimization.
\newblock In \emph{International Conference on Artificial Intelligence and
  Statistics}, 2021.

\bibitem[Zhang et~al.(2019{\natexlab{a}})Zhang, Li, Nado, Martens, Sachdeva,
  Dahl, Shallue, and Grosse]{zhang2019algorithmic}
Guodong Zhang, Lala Li, Zachary Nado, James Martens, Sushant Sachdeva, George
  Dahl, Chris Shallue, and Roger~B Grosse.
\newblock Which algorithmic choices matter at which batch sizes? insights from
  a noisy quadratic model.
\newblock \emph{Advances in neural information processing systems},
  32:\penalty0 8196--8207, 2019{\natexlab{a}}.

\bibitem[Zhang et~al.(2021)Zhang, Bao, Lessard, and Grosse]{zhang2020unified}
Guodong Zhang, Xuchan Bao, Laurent Lessard, and Roger Grosse.
\newblock A unified analysis of first-order methods for smooth games via
  integral quadratic constraints.
\newblock \emph{Journal of Machine Learning Research}, 22:\penalty0 1--39,
  2021.

\bibitem[Zhang and Yu(2020)]{zhang2019convergence}
Guojun Zhang and Yaoliang Yu.
\newblock Convergence of gradient methods on bilinear zero-sum games.
\newblock In \emph{International Conference on Learning Representations}, 2020.

\bibitem[Zhang et~al.(2019{\natexlab{b}})Zhang, Hong, and
  Zhang]{zhang2019lower}
Junyu Zhang, Mingyi Hong, and Shuzhong Zhang.
\newblock On lower iteration complexity bounds for the saddle point problems.
\newblock \emph{arXiv preprint arXiv:1912.07481}, 2019{\natexlab{b}}.

\end{thebibliography}
\bibliographystyle{plainnat}

\newpage
\appendix
\onecolumn

\section{Technical Proofs}
For notational convenience, we define the gradient vector field of minimax games $V(\bz) = [\nabla_\bx f(\bx, \by)^\top, -\nabla_\by f(\bx, \by)^\top]^\top$ and its associated Jacobian matrix at Nash equilibrium $\bz^*$:
\begin{equation}\label{eq:jacobian}
    \jacobian = \bmat{\nabla_\bx^2 f(\bx^*, \by^*) & \nabla_{\bx\by}^2 f(\bx^*, \by^*) \\ -\nabla_{\by\bx}^2 f(\bx^*, \by^*) & -\nabla_\by^2 f(\bx^*, \by^*)} \triangleq \bmat{\bA & \bB \\ - \bB^\top & \bC}
\end{equation}

\subsection{Proof of Theorem~\ref{thm:alt-gda-bilinear}}
For the bilinear games, the spectral radius of $\nabla F_\eta^\textup{Alt}$ is easy to bound. In particular, we have
\begin{equation}
    \nabla F_\eta^\textup{Alt} = \bmat{\iden & -\eta \bB \\ \eta \bB^\top & \iden - \eta^2 \bB^\top \bB}.
\end{equation}
Here, we define the SVD decomposition of $\bB$ as $\bB = \bU \hat{\bB} \bV^\top$ where $\hat{\bB}$ is diagonal, hence we have $\rho(\nabla F_\eta^\textup{Alt})$ equivalent to the spectral radius of the following matrix
\begin{equation}
    \bmat{\iden & -\eta \hat{\bB} \\ \eta \hat{\bB}^\top & \iden - \eta^2 \hat{\bB}^\top \hat{\bB}},
\end{equation}
whose eigenvalues satisfying $(x - 1)^2 + \eta^2 \lambda x = 0, \; \lambda \in \mathrm{Sp}(\bB^\top \bB)$. As long as $|2 - \eta^2 \lambda| \leq 2$, we have the roots of this polynomial satisfying $|x| = 1$. Therefore, we have $\rho(\nabla F_\eta^\textup{Alt}) = 1$ as long as $\eta \leq \frac{2}{\sigma_\text{max}(\bB)}$. In the meantime, it can shown that $\rho(\nabla F_\eta^\textup{Alt})$ is diagonalizable, so by \cite[Lemma 3]{gidel2019negative}, the iterates of Alt-GDA stay bounded. This finishes the proof.

\subsection{Proof of Theorem~\ref{thm:sim-gda}}\label{app:B}
To prove Theorem~\ref{thm:sim-gda}, we first claim that all eigenvalues of $\jacobian$ in \eqref{eq:jacobian} fall within the following set:
\begin{equation}
    \mathcal{K} = \left\{ \lambda \in \comp: |\lambda| \leq \sqrt{2}L, \realp \lambda \geq \mu > 0 \right\}.
\end{equation}
We first prove $\realp \lambda \geq \mu$. Let $\lambda \triangleq a + bi$ be a complex eigenvalue of $\jacobian$ such that $\jacobian \bv = \lambda \bv$. In general, the eigenvector $\bv$ is a complex vector and we let $\bv = \bu + \bw i$. Then, one can show
\begin{equation}
    \realp(\lambda) = \frac{\bu^\top \jacobian\bu + \bw^\top \jacobian \bw}{\bu^\top \bu + \bw^\top\bw} = \frac{\bu_1^\top \bA \bu_1 + \bu_2^\top \bC \bu_2 + \bw_1^\top\bA\bw_1 + \bw_2^\top \bC \bw_2}{\bu^\top \bu + \bw^\top\bw}
\end{equation}
where $\bu_1 \in \real^m$ is the first half of the vector $\bu$ and $\bu_2 \in \real^n$ is the second half (the same for $\bw_1, \bw_2$). As we know that $\bA \succeq \mu_\bx \iden \succeq \mu \iden$ and $\bC \succeq \mu_\by \iden \succeq \mu \iden$, we have
\begin{equation}
    \realp(\lambda) \geq \frac{\mu \bu_1^\top\bu_1 + \mu \bu_2^\top\bu_2 + \mu \bw_1^\top\bw_1 + \mu \bw_2^\top\bw_2}{\bu^\top \bu + \bw^\top\bw} = \mu
\end{equation}
We next prove $|\lambda| \leq \sqrt{2}L$. To this end, it suffices to show $\lambda_\text{max}(\jacobian^\top \jacobian) \leq 2L^2$. Recall the definition of $\jacobian$ in \eqref{eq:jacobian}, we have
\begin{equation}
    \jacobian^\top \jacobian = \bmat{\bA & -\bB \\ \bB^\top & \bC} \bmat{\bA & \bB \\ -\bB^\top & \bC} = \bmat{\bA^2 + \bB\bB^\top & \bA\bB - \bB\bC \\ \bB^\top \bA - \bC\bB^\top & \bB^\top\bB + \bC^2}.
\end{equation}
Hence,
\begin{equation}
    \lambda_\text{max}\left(\jacobian^\top \jacobian\right) = \max_{\|\bv\|=1} \bv^\top \jacobian^\top \jacobian\bv = \max_{\|\bv\|=1} \bv_1^\top (\bA^2 + \bB\bB^\top) \bv_1 + \bv_2^\top (\bC^2 + \bB^\top \bB) \bv_2
\end{equation}
Because we assume $\bA \preceq L_\bx \iden \preceq L \iden$, $\bC \preceq L_\by \iden \preceq L \iden$ and $\|\bB\|_2 \leq L_{\bx\by} \leq L$, we have
\begin{equation}\label{eq:jacobian-upper}
    \lambda_\text{max}\left(\jacobian^\top \jacobian\right) \leq \max_{\|\bv\|=1} 2L^2 (\bv_1^\top \bv_1 + \bv_2^\top \bv_2) = 2L^2.
\end{equation}
Therefore, we get $|\lambda| \leq \sqrt{2}L$. Now the convergence rate bound of Sim-GDA reduces to the following problem:
\begin{equation}
    \min_\eta \max_{\lambda \in \mathcal{K}} |1 - \eta \lambda| = \min_\eta \max_{\lambda \in \mathcal{K}} \sqrt{(1 - \eta \realp(\lambda))^2 + \eta^2 \imag(\lambda)^2}
\end{equation}
where the maximum modulus is achieved at the point $\lambda = \mu + \sqrt{2L^2 - \mu^2}i$. Hence, we have
\begin{equation}\label{eq:spectral-radius-thm4-fixed}
    \rho(\nabla F_\eta^\textup{Sim}(\bz^*)) \leq  \min_\eta \max_{\lambda \in \mathcal{K}} |1 - \eta \lambda| = \min_\eta \sqrt{1 - 2\eta \mu + 2\eta^2 L^2} = \sqrt{1 - \frac{\mu^2}{2L^2}}.
\end{equation}
By invoking Theorem~\ref{thm:local}, we finish our proof.

\subsection{Proof of Theorem~\ref{thm:alt-gda}}
Recall the updates of Alt-GDA:
\begin{equation*}
    \bx_{t+1} = \bx_t - \eta \nabla_\bx f(\bx_t, \by_t), \; \by_{t+1} = \by_t + \eta \nabla_\by f({\color{red}\bx_{t+1}}, \by_t).
\end{equation*}
By definition, we have the Jacobian matrix of Alt-GDA updates in the following form:
\begin{equation}\label{eq:joint-jacobian-alt}
    \nabla F_\eta^\textup{Alt}(\bz^*) = \begin{bmatrix} \iden & \zero \\ \eta \bB^\top & \iden - \eta \bC \end{bmatrix}
    \begin{bmatrix} \iden - \eta \bA & -\eta \bB \\ \zero & \iden \end{bmatrix} 
    = \begin{bmatrix} \iden - \eta \bA & -\eta \bB \\ \eta \bB^\top (\iden - \eta \bA) & \iden - \eta \bC - \eta^2 \bB^\top \bB \end{bmatrix}.
\end{equation}
Without loss of generality, we assume matrices $\bA$ and $\bC$ to be diagonal with eigenvalues $[\alpha]_i$ and $[\beta]_j$. We have the characteristic polynomial:
\begin{multline}
    \mathrm{Det}(\lambda \iden - \nabla F_\eta^\textup{Alt} (\bz^*)) = \mathrm{Det}(\lambda \iden - (\iden - \eta \bA)) \\ \mathrm{Det}(\lambda \iden  - (\iden - \eta \bC - \eta^2 \bB^\top \bB - \eta^2\bB^\top (\iden - \eta \bA)(\lambda\iden - \iden + \eta\bA)^{-1}\bB)),
\end{multline}
where we used properties of the Schur complement. For $\mathrm{Det}(\lambda \iden - \nabla F_\eta^\textup{Alt}(\bz^*))$ to be zero, $\lambda$ has to one of the eigenvalues of the following matrix:
\begin{equation}
    \bM \triangleq \iden - \eta \bC - \eta^2 \bB^\top \mathrm{Diag}\left(\frac{\lambda}{\lambda - 1 + \eta \alpha_i}\right) \bB.
\end{equation}
\textbf{In the case of $\lambda$ being real}, it is easy to show that for any $\eta \leq \frac{1}{2L}$, we have
\begin{equation}
    |\lambda| \leq \max\{1 - \eta \alpha_{\text{min}}, 1 - \eta \beta_{\text{min}}\}.
\end{equation}
We prove that by contradiction. Suppose $\lambda > \max\{1 - \eta \alpha_{\text{min}}, 1 - \eta \beta_{\text{min}}\} \geq 0$, then we have 
\begin{equation}
    \lambda_\text{max}(\bM) \leq 1 - \eta \lambda_\text{min}(\bC) = 1 - \eta \beta_\text{min}.
\end{equation}
That is because the term $\eta^2 \bB^\top \mathrm{Diag}\left(\frac{\lambda}{\lambda - 1 + \eta \alpha_i}\right) \bB$ is positive semi-definite when $\lambda > \max\{1 - \eta \alpha_{\text{min}}, 1 - \eta \beta_{\text{min}}\} \geq 0$. Since we know $\lambda$ is one of the eigenvalue of $\bM$ and hence it has to be smaller than $1 - \eta \beta_\text{min}$, contradiction.

Suppose $\lambda < -\max\{1 - \eta \alpha_{\text{min}}, 1 - \eta \beta_{\text{min}}\} \leq 0$, we have
\begin{equation}
    \bM \succeq \iden - \eta \bC - \eta^2 \bB^\top \bB.
\end{equation}
For bilinear games where $\bC = \zero$, we have $\bM \succeq \zero$ and hence $\lambda$ is impossible to be smaller than $-1$. On the other hand, since we have $\eta \leq \frac{1}{2L}$, we know $\bM \succeq \zero$, contradiction again. Therefore, we proved that $|\lambda| \leq \max\{1 - \eta \alpha_{\text{min}}, 1 - \eta \beta_{\text{min}}\}$.

\textbf{In the case of $\lambda$ being complex}, we let $\lambda = a + b i$ with $b \neq 0$ and $\bv$ be the eigenvector associated with $\lambda$ such that $\bM \bv = \lambda \bv$. Then we have the following identities:
\begin{equation}
\begin{aligned}
    \bv^{\hermconj} (\bM + \bM^{\hermconj}) \bv &= 2 \realp(\lambda) = 2a, \\
    \bv^{\hermconj} (\bM - \bM^{\hermconj}) \bv &= 2 \imag(\lambda) i = 2b i.
\end{aligned}
\end{equation}
Plugging the value of $\bM$, we have
\begin{equation}\label{eq:real-identity}
a = \frac{1}{2}\bv^{\hermconj} (\bM + \bM^{\hermconj}) \bv = \sum_j (1 - \eta \beta_j)|\bv_j|^2 - \eta^2 \sum_j  |(\bB\bv)_j|^2 \frac{a^2 - a(1 - \eta \alpha_j) + b^2}{(a - 1 + \eta\alpha_j)^2 + b^2}
\end{equation}
and
\begin{equation}\label{eq:imag-identity}
bi = \frac{1}{2}\bv^{\hermconj} (\bM - \bM^{\hermconj}) \bv = \eta^2 \sum_j |(\bB \bv)_j|^2 \frac{(1 - \eta \alpha_j) b i}{(a - 1 + \eta\alpha_j)^2 + b^2}.
\end{equation}
From \eqref{eq:imag-identity}, one can get
\begin{equation}\label{eq:imag-identity-2}
    \eta^2 \sum_j |(\bB \bv)_j|^2 \frac{(1 - \eta \alpha_j) }{(a - 1 + \eta\alpha_j)^2 + b^2} = 1.
\end{equation}
Next, combining \eqref{eq:real-identity} and \eqref{eq:imag-identity-2}, we have
\begin{equation}\label{eq:comb-identity}
    \sum_j (1 - \eta \beta_j)|\bv_j|^2 - \eta^2 |(\bB\bv)_j|^2 \frac{a^2 + b^2}{\Delta_j} = 0,
\end{equation}
where $\Delta_j = (a - 1 + \eta\alpha_j)^2 + b^2$. It follows from \eqref{eq:comb-identity} and \eqref{eq:imag-identity-2} that
\begin{equation}
\begin{aligned}
    1 &= \eta^2 \sum_j |(\bB \bv)_j|^2 \frac{(1 - \eta \alpha_j) }{\Delta_j} \leq \eta^2 \sum_j |(\bB \bv)_j|^2 \frac{(1 - \eta \alpha_\text{min}) }{\Delta_j} \\
    &= \eta^2 \frac{1 - \eta \alpha_\text{min}}{a^2 + b^2}\sum_j |(\bB \bv)_j|^2 \frac{a^2 + b^2 }{\Delta_j} \\
    \overset{\eqref{eq:comb-identity}}&{=} \frac{1 - \eta \alpha_\text{min}}{a^2 + b^2}\sum_j (1 - \eta \beta_j)|\bv_j|^2  \leq \frac{(1 - \eta \alpha_\text{min})(1 - \eta \beta_\text{min})}{a^2 + b^2}
\end{aligned}.    
\end{equation}
As a result,
\begin{equation}\label{eq:complex-bound}
    |\lambda|^2 = a^2 + b^2 \leq (1 - \eta \alpha_\text{min})(1 - \eta \beta_\text{min}).
\end{equation}

\subsection{Proof of Theorem~\ref{prop:sim-gda}}
Recall that the Jacobian matrix $\nabla F_\eta^\textup{Sim}(\bz^*)$ of Sim-GDA:
\begin{equation}
    \nabla F_\eta^\textup{Sim}(\bz^*) = \bmat{\iden - \eta \bA & -\eta \bB \\ \eta \bB^\top & \iden - \eta \bC}.
\end{equation}
To bound its spectral radius, we first compute its characteristic polynomial and simplify it with the Schur complement.
\begin{equation}
    \mathrm{Det} \left(\lambda\iden - \nabla F_\eta^\textup{Sim}(\bz^*)\right) = \mathrm{Det}(\lambda \iden - (\iden - \eta \bC)) \mathrm{Det}(\lambda \iden - (\iden - \eta \bA - \eta^2\bB (\lambda \iden - (\iden - \eta\bC))^{-1}\bB^\top))
\end{equation}
\textbf{In the case of $\lambda$ being real}, for $\eta \leq \frac{1}{L}$, one can prove that $\lambda$ is within the range $(0, 1)$ by contradiction argument. Without loss of generality, we assume matrices $\bA$ and $\bC$ to be diagonal with eigenvalues $[\alpha]_i$ and $[\beta]_j$. 
Let assume $\lambda > 1 - \eta \beta_\text{min}$, then we claim that $\lambda \leq 1 - \frac{\eta}{L}\lambda_\text{min}(\bB \bB^\top)$. The key is (by $\lambda < 1$)
\begin{equation}
    (\lambda \iden - (\iden - \eta \bC))^{-1} \succeq (\eta\bC)^{-1} \succeq \frac{1}{\eta L_\by} \iden \succeq \frac{1}{\eta L }\iden
\end{equation}
Therefore, we have a upper bound for $\bM \triangleq \iden - \eta \bA - \eta^2\bB (\lambda \iden - (\iden - \eta\bC))^{-1}\bB^\top)$
\begin{equation}
    \bM \preceq \iden - \frac{\eta}{L}\bB \bB^\top \preceq \left(1 - \frac{\eta}{L}\lambda_\text{min}(\bB \bB^\top)\right) \iden.
\end{equation}
Hence, $\lambda$, as one of the eigenvalues of $\bM$, has to be smaller than $1 - \frac{\eta}{L}\lambda_\text{min}(\bB \bB^\top)$. In summary, we proved that
\begin{equation}
    0 < \lambda \leq \max\left\{1 - \eta \beta_\text{min}, 1 - \frac{\eta}{L}\lambda_\text{min}(\bB \bB^\top) \right\}
\end{equation}

\textbf{In the case of $\lambda$ being complex}, we claim that $\realp(\lambda) \leq 1 - \frac{\eta}{2}\mu_\by$. Let $\lambda = a + b i$ with $b \neq 0$ and $\bv$ be the eigenvector associated with $\lambda$ such that $\bM \bv = \lambda \bv$. Then we have the following identities:
\begin{equation}
\begin{aligned}
    \bv^{\hermconj} (\bM + \bM^{\hermconj}) \bv &= 2 \realp(\lambda) = 2a \\
    \bv^{\hermconj} (\bM - \bM^{\hermconj}) \bv &= 2 \imag(\lambda) i = 2b i
\end{aligned}
\end{equation}
Plugging the value of $\bM$, we have
\begin{equation}\label{eq:no-strong-convex-identity-2}
a = \frac{1}{2}\bv^{\hermconj} (\bM + \bM^{\hermconj}) \bv = \sum_j (1 - \eta \alpha_j)|\bv_j|^2 - \eta^2 \sum_j  |(\bB^\top\bv)_j|^2 \frac{a - (1 - \eta \beta_j)}{(a - 1 + \eta\beta_j)^2 + b^2}
\end{equation}
and
\begin{equation}\label{eq:no-strong-convex-identity}
bi = \frac{1}{2}\bv^{\hermconj} (\bM - \bM^{\hermconj}) \bv = \eta^2 \sum_j |(\bB^\top \bv)_j|^2 \frac{bi}{(a - 1 + \eta\beta_j)^2 + b^2}
\end{equation}
From \eqref{eq:no-strong-convex-identity}, we get
\begin{equation}
    \eta^2 \sum_j |(\bB^\top \bv)_j|^2 \frac{1}{(a - 1 + \eta\beta_j)^2 + b^2} = 1
\end{equation}
Next, combining \eqref{eq:no-strong-convex-identity} and \eqref{eq:no-strong-convex-identity-2}, we have
\begin{equation}
\begin{aligned}
    2a - 1 &=  \sum_j (1 - \eta \alpha_j)|\bv_j|^2 - \eta^2 \sum_j  |(\bB^\top\bv)_j|^2 \frac{\eta \beta_j}{(a - 1 + \eta\beta_j)^2 + b^2} \\
    &\leq \sum_j (1 - \eta \alpha_\text{min})|\bv_j|^2 - \eta^2 \sum_j  |(\bB^\top\bv)_j|^2 \frac{\eta \beta_\text{min}}{(a - 1 + \eta\beta_j)^2 + b^2} \\ \overset{\eqref{eq:no-strong-convex-identity}}&{=} 1 - \eta \beta_\text{min}
\end{aligned}
\end{equation}
Therefore, we proved our claim that $\realp(\lambda) = a \leq 1 - \frac{\eta}{2}\beta_\text{min}$.

Then, we could prove $\jacobian = \frac{1}{\eta}(\iden - \nabla F_\eta^\textup{Sim}(\bz^*))$ has the operator norm $\|\jacobian\| \leq \sqrt{2}L$ by the same argument in \eqref{eq:jacobian-upper}. Consequently, we have $|\imag(\lambda)| \leq \eta\sqrt{2L^2 - \frac{1}{4}\mu_\by^2}$. Follows immediately, we have
\begin{equation}
    |\lambda| = \sqrt{\realp(\lambda)^2 + \imag(\lambda)^2} \leq (1 - \frac{\eta}{2} \mu_\by)^2 + \eta^2 (2L^2 - \frac{1}{4}\mu_\by^2) = 1 - \eta\mu_\by + 2\eta^2 L^2
\end{equation}

\subsection{Proof of Theorem~\ref{prop:alt-gda}}
As in the proof of Theorem~\ref{thm:alt-gda}, we analyze the eigenvalues of $\nabla F_\eta^\textup{Alt}(\bz^*)$. Recall $\nabla F_\eta^\textup{Alt}(\bz^*)$ has the following form:
\begin{equation}\label{eq:joint-jacobian-alt-new}
    \nabla F_\eta^\textup{Alt}(\bz^*) = 
    \begin{bmatrix} \iden - \eta \bA & -\eta \bB \\ \zero & \iden \end{bmatrix} \begin{bmatrix} \iden & \zero \\ \eta \bB^\top & \iden - \eta \bC \end{bmatrix}
    = \begin{bmatrix} \iden - \eta \bA - \eta^2 \bB\bB^\top & -\eta \bB(\iden - \eta \bC) \\ \eta \bB^\top & \iden - \eta \bC \end{bmatrix}
\end{equation}
Notice that this matrix in \eqref{eq:joint-jacobian-alt-new} is slightly different from the one defined in \eqref{eq:joint-jacobian-alt}, but they have the same eigen-spectrum.
Without loss of generality, we assume matrices $\bA$ and $\bC$ to be diagonal with eigenvalues $[\alpha]_i$ and $[\beta]_j$. We then have the characteristic polynomial:
\begin{multline}
    \mathrm{Det}(\lambda \iden - \nabla F_\eta^\textup{Alt}(\bz^*)) = \mathrm{Det}(\lambda \iden - (\iden - \eta \bC)) \\ \mathrm{Det}(\lambda \iden - (\iden - \eta \bA - \eta^2 \bB \bB^\top - \eta^2\bB (\iden - \eta \bC)(\lambda\iden - \iden + \eta\bC)^{-1}\bB^\top))
\end{multline}
where we used Schur complement. For $\mathrm{Det}(\lambda \iden - \nabla F_\eta^\textup{Alt}(\bz^*))$ to be zero, $\lambda$ has to one of the eigenvalues of the following matrix:
\begin{equation}
    \bM \triangleq \iden - \eta \bA - \eta^2 \bB \bB^\top - \eta^2\bB (\iden - \eta \bC)(\lambda\iden - \iden + \eta\bC)^{-1}\bB^\top
\end{equation}
\textbf{In the case of $\lambda$ being real}, for any $\eta \leq \frac{1}{2L}$, it is easy to show that $\lambda$, as one of the eigenvalues of $\bM$, is within $(0, 1)$. Let us first assume $\lambda > 1 - \eta \beta_\text{min}$, we have $\bB (\iden - \eta \bC)(\lambda\iden - \iden + \eta\bC)^{-1}\bB^\top \succeq \zero$. Hence, we have
\begin{equation}
    \bM \preceq \iden - \eta \bA - \eta^2 \bB\bB^\top \preceq \iden - \eta^2 \bB\bB^\top \preceq (1 - \eta^2 \mu_{\bx\by}^2) \iden
\end{equation}
As a consequence, we know $|\lambda| \leq 1 - \eta^2 \mu_{\bx\by}^2$.

\textbf{In the case of $\lambda$ being complex}, we could reuse the result of \eqref{eq:complex-bound}
\begin{equation}
    |\lambda| \leq \sqrt{(1 - \eta \alpha_\text{min})(1 - \eta \beta_\text{min})} \leq \sqrt{1 - \eta \mu_\by}
\end{equation}

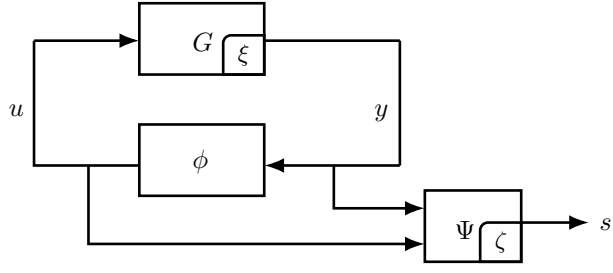
\begin{figure}[t]
\centering
\begin{tikzpicture}[x=0.75pt,y=0.75pt,yscale=-0.9,xscale=0.9]
\draw   (100,105) -- (170,105) -- (170,145) -- (100,145) -- cycle ;
\draw   (100,173) -- (170,173) -- (170,213) -- (100,213) -- cycle ;
\draw   (170,126) -- (246,126) ;
\draw    (246,126) -- (246,196) ;
\draw[-{Latex[scale=1.0]}]    (246,196) -- (170,196) ;
\draw    (100,196) -- (41,196) -- (41,126) ;
\draw[-{Latex[scale=1.0]}]    (41,126) -- (100,126) ;
\draw   (147,127.4) .. controls (147,124.97) and (148.97,123) .. (151.4,123) -- (170,123) -- (170,145) -- (147,145) -- cycle ;
\draw[-{Latex[scale=1.0]}]   (209,196) -- (209,220) -- (260,220) ;
\draw[-{Latex[scale=1.0]}]    (71.5,196) -- (71.5,240) -- (260,240) ;
\draw   (260,210) -- (314,210) -- (314,250) -- (260,250) -- cycle ;
\draw   (291,232.4) .. controls (291,229.97) and (292.97,228) .. (295.4,228) -- (314,228) -- (314,250) -- (291,250) -- cycle ;
\draw[-{Latex[scale=1.0]}]    (314,228) -- (352,228) ;

\draw (128,120) node [anchor=north west][inner sep=0.75pt]    {$G$};
\draw (128,185) node [anchor=north west][inner sep=0.75pt]    {$\phi$};
\draw (230,160) node [anchor=north west][inner sep=0.75pt]    {$y$};
\draw (25,160) node [anchor=north west][inner sep=0.75pt]    {$u$};
\draw (153.4,126.4) node [anchor=north west][inner sep=0.75pt]  [font=\small]  {$\xi $};
\draw (275,225) node [anchor=north west][inner sep=0.75pt]    {$\Psi $};
\draw (297.4,231.4) node [anchor=north west][inner sep=0.75pt]  [font=\small]  {$\zeta $};
\draw (356,224) node [anchor=north west][inner sep=0.75pt]    {$s$};
\end{tikzpicture}
\caption{Feedback interconnection between a system $G$ (optimization algorithm) with state matrices $(A, B, C, D)$ and a nonlinearity $\phi$. An IQC is a constraint on $(y, u)$ satisfied by $\phi$.}
\label{fig:my_label}
\end{figure}

\section{Details about IQC framework}\label{app:iqc}
Borrowing the notations from~\cite{lessard2016analysis}, we frame various first-order algorithms as a unified linear dynamical system\footnote{This linear dynamical system can represent any first-order methods.} in feedback with a nonlinearity $\phi: \real^d \rightarrow \real^d$,
\begin{equation}\label{eq:dynamical-system}
\begin{aligned}
    \xi_{t+1} &= A \xi_{t} + B u_t \\
    y_t &= C \xi_t + D u_t \\
    u_t &= \phi(y_t).
\end{aligned}
\end{equation}
At each iteration $t = 0, 1, ...$, $u_t \in \real^d$ is the control input, $y_t \in \real^d$ is the output, and $\xi_t \in \real^{nd}$ is the state for algorithms with $n$ step of memory. 
The state matrices $A, B, C, D$ differ for various algorithms. For most algorithms we consider in the paper, they have the general form:
\begin{equation*}
    \left[ 
    \begin{array}{c|c}
        A & B \\ \hline
        C & D
    \end{array}
    \right]
    = 
    \left[ 
    \begin{array}{cc|c}
        (1 + \beta) \iden_d & -\beta \iden_d & -\eta \iden_d \\ 
        \iden_d & \zero_d & \zero_d \\ \hline
        (1 + \alpha)\iden_d & -\alpha \iden_d & \zero_d
    \end{array}
    \right],
\end{equation*}
where $\iden_d$ and $\zero_d$ are the identity and zero matrix of size $d \times d$, respectively.
Often, the nonlinear function $\phi$ is the troublesome function we wish to analyze. Although we do not know $\phi$ exactly, we assume to have some knowledge of the constraints it imposes on the input-output pair $(y, u)$. For example, we may assume $\phi$ to be $L$-Lipschitz, which implies
$\|u_t - u^*\|_2 \leq L \|y_t - y^* \|_2 $ for all $t$ with $u^* = \phi(y^*)$ as a fixed point. In matrix form, this is
\begin{equation}\label{eq:lipschitz-quad-constraint}
    \begin{bmatrix}
    y_t - y^* \\
    u_t - u^*
    \end{bmatrix}^\top
    \begin{bmatrix}
    L^2 \iden_d & \zero_d \\
    \zero_d & - \iden_d
    \end{bmatrix}
    \begin{bmatrix}
    y_t - y^* \\
    u_t - u^*
    \end{bmatrix} \geq 0.
\end{equation}
We can also characterize strong convexity of $f$ and $g$ by similar quadratic constraints. Notably, the above constraint is very special in that it only manifests itself as separate quadratic constraints on each $(y_t, u_t)$. It is possible to specify quadratic constraints that couple different $t$ values. To achieve that, we follow \cite{lessard2016analysis} and adopt auxiliary sequences $\zeta, s$ together with a map $\Psi$ characterized by matrices $(A_\Psi, B_\Psi^y, B_\Psi^u, C_\Psi, D_\Psi^y, D_\Psi^u)$:
\begin{equation}\label{eq:ass-ref}
\begin{aligned}
    \zeta_{t+1} &= A_\Psi \zeta_t + B_\Psi^y y_t + B_\Psi^u u_t, \\
    s_{t} &= C_\Psi \zeta_t + D_\Psi^y y_t + D_\Psi^u u_t.
\end{aligned}
\end{equation}
The equations~\eqref{eq:ass-ref} define an affine map $s = \Psi(y, u)$, where $s_t$ could be a function of all past $y_i$ and $u_i$ with $i \leq t$.
We consider the quadratic form $(s_t - s^*)^\top M (s_t-s^*)$ for a given matrix $M$ with $s^*$ and $\xi^*$ fixed points of \eqref{eq:ass-ref}. We note that the quadratic form is a function of $(y_0,\dots, y_t, u_0,\dots, u_t)$ that is determined by our choice of $(\Psi, M)$. In particular, we can recover constraint~\eqref{eq:lipschitz-quad-constraint} with
\begin{equation}\label{eq:psi}
    \Psi = \left[ 
    \begin{array}{c|cc}
        A_\Psi & B_\Psi^y & B_\Psi^u \\\hline
        C_\Psi & D_\Psi^y & D_\Psi^u \\ 
    \end{array}
    \right]
    = 
    \left[ 
    \begin{array}{c|cc}
        \zero_d & \zero_d & \zero_d \\\hline
        \zero_d & \iden_d & \zero_d \\
        \zero_d & \zero_d & \iden_d \\ 
    \end{array}
    \right], \qquad
    M = \begin{bmatrix}
    L^2 \iden_d & \zero_d \\
    \zero_d & - \iden_d
    \end{bmatrix}.
\end{equation}
Combining the dynamics~\eqref{eq:dynamical-system} with the map $\Psi$ (by eliminating $y_t$), we obtain
\begin{equation}\label{eq:dynamics-comb}
\begin{aligned}
    \bmat{\xi_{t+1} \\ \zeta_{t+1}} &= \bmat{A & 0 \\ B_{\Psi}^y C & A_\Psi} \bmat{\xi_{t} \\ \zeta_{t}} + \bmat{B \\ B_\Psi^u + B_\Psi^y D} u_t, \\
    s_t &= \bmat{D_{\Psi}^y C & C_{\Psi}} \bmat{\xi_{t} \\ \zeta_{t}} + \bmat{D_\Psi^u + D_\Psi^y D} u_t.
\end{aligned}
\end{equation}
More succinctly, \eqref{eq:dynamics-comb} can be written as
\begin{equation}\label{eq:dynamics-compact}
\begin{aligned}
    x_{t+1} &= \hat{A} x_t + \hat{B} u_t \\
    s_t &= \hat{C}x_t + \hat{D} u_t
\end{aligned},
\qquad \text{where } x_t \triangleq \begin{bmatrix} \xi_t \\ \zeta_t \end{bmatrix}.
\end{equation}
With these definitions in hand, we now state the main result of verifying exponential convergence. Basically, we build a Linear Matrix Inequality (LMI) to guide the search for the parameters of a quadratic Lyapunov function in order to establish a rate bound.
\begin{theorem}[\cite{zhang2020unified}]
\label{thm:main-thm}
Consider the dynamical system~\eqref{eq:dynamical-system}.
Suppose the vector field $F$ satisfies the IQC $(\Psi, M)$ and define $(\hat A, \hat B, \hat C, \hat D)$ according to~\eqref{eq:psi}--\eqref{eq:dynamics-compact}, we have the following linear matrix inequality (LMI):
    \begin{equation}\label{eq:sdp}
        \begin{bmatrix}
        \hat{A}^\top P \hat{A} - \rho^2 P & \hat{A}^\top P \hat{B}\\
        \hat{B}^\top P \hat{A} & \hat{B}^\top P \hat{B}
        \end{bmatrix} +
        \lambda \begin{bmatrix} \hat{C} & \hat{D} \end{bmatrix}^\top M \begin{bmatrix} \hat{C} & \hat{D} \end{bmatrix} \preceq 0.
    \end{equation}
    If this LMI is feasible for some $P \succ 0$, $\lambda \geq 0$ and $\rho > 0$, we have
    \begin{equation}
        (x_{t+1} - x^*)^\top (P \otimes \iden_d) (x_{t+1} - x^*)  \leq \rho^2 (x_t - x^*)^\top (P \otimes \iden_d) (x_t - x^*).
    \end{equation}
    Consequently, for any $\xi_0$ and $\zeta_0 = \zeta^*$, we obtain
    \begin{equation}
        \|\xi_t - \xi^*\|_2^2 \leq \mathrm{cond}(P) \rho^{2t} \|\xi_0 - \xi^*  \|_2^2.
    \end{equation}
\end{theorem}
\begin{rem}
The LMI~\eqref{eq:sdp} can be extended to the case of multiple constraints with $(\Psi_i, M_i)$ (see~\cite[Page 12]{lessard2016analysis} for details).
\end{rem}
To apply Theorem~\ref{thm:main-thm}, we seek to solve the semidefinite program (SDP) of finding the minimal $\rho$ such that the LMI~\eqref{eq:sdp} is feasible. For simple algorithms, one can typically solve the SDP analytically. Nevertheless, one may only get a numerical proof when the algorithm of interest is complicated and the resulting SDP is hard to solve.

\subsection{Analyzing Alt-GDA with IQC framework}
Recall that we are concerned with the bilinear saddle point problem:
\begin{equation}
    \min_{\bx\in \real^m} \max_{\by \in \real^n} f(\bx) + \bx^\top \bB \by - g(\by).
\end{equation}
For Alt-GDA, it has the following update rule:
\begin{equation}
\begin{aligned}
    \bx_{t+1} &= \bx_t - \eta \bB \by_t - \eta \nabla_\bx f(\bx_t) \\
    \by_{t+1} &= \by_t + \eta \bB^\top \bx_{t+1} - \eta \nabla_\by g(\by_t)
\end{aligned}
\end{equation}
We can frame it as a linear dynamical system in feedback with the state matrices:
\begin{equation}\label{eq:state-matrices}
    \left[ 
    \begin{array}{c|c}
        A & B \\ \hline
        C & D
    \end{array}
    \right]
    = 
    \left[ 
    \begin{array}{cc|cc}
        \iden_m & -\eta \bB & -\eta \iden_m & \zero_{m\times n} \\ 
        \eta \bB^\top & \iden_n - \eta^2 \bB^\top \bB & -\eta^2 \bB^\top & - \eta \iden_n \\ \hline
        \iden_m & \zero_{m\times n} & \zero_m & \zero_{m\times n} \\ 
        \zero_{n\times m} & \iden_n & \zero_{n \times m} & \zero_n
    \end{array}
    \right],
\end{equation}
In our case, we have $\xi_t = y_t = [\bx_t^\top, \by_t^\top]^\top$ and $u_t = [\nabla_\bx f(\bx), -\nabla_\by g(\by)]$. Further, we use the weighted off-by-one IQC defined in~\cite{lessard2016analysis} with the following representation (let $d = m + n$):
\begin{equation}\label{eq:aux-matrices}
    \Psi = \left[ 
    \begin{array}{c|cc}
        A_\Psi & B_\Psi^y & B_\Psi^u \\\hline
        C_\Psi & D_\Psi^y & D_\Psi^u \\ 
    \end{array}
    \right]
    = 
    \left[ 
    \begin{array}{c|cc}
        \zero_d & -L\iden_d & \iden_d \\\hline
        \rho^2\iden_d & L\iden_d & -\iden_d \\
        \zero_d & -\mu\iden_d & \iden_d \\ 
    \end{array}
    \right],
\end{equation}
and 
\begin{equation}\label{eq:consts}
    M_1 = \bmat{\zero_m & \zero_{m\times n} & \iden_m & \zero_{m\times n} \\
    \zero_{n\times m} & \zero_n & \zero_{n \times m} & \zero_n \\
    \iden_m & \zero_{m\times n} & \zero_m & \zero_{m\times n} \\
    \zero_{n\times m} & \zero_n & \zero_{n \times m} & \zero_n},
    \quad
    M_2 = \bmat{\zero_m & \zero_{m\times n} & \zero_m & \zero_{m\times n} \\
    \zero_{n\times m} & \zero_n & \zero_{n \times m} & \iden_n \\
    \zero_m & \zero_{m\times n} & \zero_m & \zero_{m\times n} \\
    \zero_{n\times m} & \iden_n & \zero_{n \times m} & \zero_n}.
\end{equation}
With all these matrices defined, we can solve the SDP problem~\eqref{eq:sdp} with bisection search on $\rho$. However, we can only solve SDPs with relatively small $m$ and $n$ in practice. 
Towards this end, we prove that we can reduce any problem of \eqref{eq:bilinear-sp} to the case with $m = n = 1$, which is numerically easy to solve. First, we assume without loss of generality that $m \leq n$ and $\bB \in \real^{m \times n}$ is diagonal. In the case that $\bB$ is not diagonal, we can do singular value decomposition to get $\bB = \bU \hat{\bB} \bV^\top$ where $\hat{\bB} \in \real^{m \times n}$ is diagonal and then absorb $\bU \in \real^{m \times m}$ and $\bV \in \real^{n \times n}$ into $\bx$ and $\by$ to get the following equivalent problem:
\begin{equation}
    \min_{\hat{\bx} \in \real^m} \max_{\hat{\by} \in \real^n} f(\bU \hat{\bx}) + \hat{\bx}^\top \hat{\bB} \hat{\by} - g(\bV\hat{\by}),
\end{equation}
where $\hat{\bx} = \bU^\top \bx$ and $\hat{\by} = \bV^\top \by$. Further, one can show $f^\prime(\bx) = f(\bU \bx)$ (or $g^\prime(\by) = g(\bV\by)$) is also $L$-smooth and $\mu$-strongly convex as $f$ (or $g$) is. This is because $\bU$ and $\bV$ are both orthogonal matrices. Therefore, we can assume $\bB$ to be diagonal without loss of generality. 
Next, we prove that the IQC-certified rate of Alt-GDA for \eqref{eq:bilinear-sp} is no worse than the rate of the same problem with $m = n = 1$. Formally, we have the following theorem.

\iqc*
\begin{proof}
For IQC, we basically search for a quadratic Lyapunov function by solving a SDP problem \eqref{eq:sdp}. By \eqref{eq:state-matrices}-\eqref{eq:consts}, we have the linear matrix inequality (LMI) as follows:
\begin{equation}\label{eq:lmi}
        \bmat{\hat{A}, \hat{B}}^\top P \bmat{\hat{A}, \hat{B}} +
         \begin{bmatrix} \hat{C} ,\hat{D} \end{bmatrix}^\top M \begin{bmatrix} \hat{C} , \hat{D} \end{bmatrix} \preceq \rho^2 \bmat{\iden , \zero }^\top P \bmat{\iden , \zero},
\end{equation}
with
\begin{gather*}
    \hat{A} = \bmat{
        \bsmat{\iden_m & -\eta \bB \\ \eta \bB^\top & \iden_n - \eta^2 \bB^\top \bB} & \zero_d \\[-1.5\medskipamount] \\
        -L \iden_d & \zero_d}
, \;
    \hat{B} = \bmat{
        \bsmat{-\eta \iden_m & \zero_{m \times n} \\ -\eta^2 \bB^\top & -\eta \iden_n} \\[-1.5\medskipamount] \\
        \iden_d},\\
    \hat{C} = \bmat{L\iden_d & \rho^2 \iden_d \\ -\mu \iden_d & \zero_d}, \; \hat{D} = \bmat{-\iden_d \\ \iden_d}, \;
    M = \bmat{\zero_m & \zero_{m\times n} & \lambda_1 \iden_m & \zero_{m\times n} \\
    \zero_{n\times m} & \zero_n & \zero_{n \times m} & \lambda_2 \iden_n \\
    \lambda_1\iden_m & \zero_{m\times n} & \zero_m & \zero_{m\times n} \\
    \zero_{n\times m} & \lambda_2 \iden_n & \zero_{n \times m} & \zero_n}.
\end{gather*}
Given that $\bB$ is diagonal, one can show that both $\bmat{\hat{A}, \hat{B}}$ and $\bmat{\hat{C}, \hat{D}}$ have very special structure. In particular, we can permute them column-wise and row-wise to get block-diagonal matrices:
\begin{equation}
    \bmat{\hat{A}, \hat{B}} = \bU 
    \underbrace{\bmat{\bsmat{1 & -\eta \bB_{11} & 0 & 0 & -\eta & 0 \\ 
                \eta \bB_{11} & 1 - \eta^2 \bB_{11}^2 & 0 & 0 & -\eta^2\bB_{11} & -\eta \\
                -L & 0 & 0 & 0 & 1 & 0 \\
                0 & -L & 0 & 0 & 0 & 1} & \cdots & \zero \\
                \vdots & \ddots & \vdots \\
                \zero & \cdots & \bsmat{1 & 0 & -\eta \\ -L & 0 & 1}}}_{\triangleq \bQ_1} \bV
\end{equation}
where both $\bU$ and $\bV$ are permutation matrices. 
In more detail, the $(1, 1)$ block is repeated $r = \min(m,n)$ times, where $\bB_{11}$ is replaced by $\bB_{ii}$, $i=1,\dots,r$ in each successive block, and the smaller block is repeated $d-r$ times.
Also we can do the same for $\bmat{\hat{C}, \hat{D}} = \bU \bQ_2 \bV$ and $\bmat{\iden, \zero} = \bU \bQ_3 \bV$. Each diagonal block of $\bQ_3$ is either $\bmat{\iden_4, \zero_{4\times 2}}$ or $\bmat{\iden_2, \zero_{2\times 1}}$. Hence, we can write \eqref{eq:lmi} in the following form:
\begin{equation}
    \bV^\top \bQ_1^\top \bU^\top P \bU \bQ_1 \bV +  \bV^\top \bQ_2^\top \bU^\top M \bU \bQ_2 \bV \preceq \rho^2 \bV^\top \bQ_3^\top \bU^\top P \bU \bQ_3 \bV,
\end{equation}
whose feasible set is equivalent to
\begin{equation}\label{eq:lmi-2}
    \bQ_1^\top \bU^\top P \bU \bQ_1 +  \bQ_2^\top \bU^\top M \bU \bQ_2 \preceq \rho^2 \bQ_3^\top \bU^\top P \bU \bQ_3.
\end{equation}
It is easy to see that $\bU^\top M \bU$ has the same block-diagonal structure as $\bQ_1$ and $\bQ_2$. If we further restrict $\bU^\top P \bU$ to have the same block-diagonal structure as $\bQ_1$ and $\bQ_2$, then it suffices to pick a $\rho$ so that each diagonal block of the LMI \eqref{eq:lmi-2} holds. Moreover, the LMI of each diagonal block is the LMI of the case $m = n = 1$, except for the last $n - m$ blocks, for which we have
\begin{equation}\label{eq:lmi-min}
    \bsmat{1 & 0 & -\eta \\ -L & 0 & 1}^\top P \bsmat{1 & 0 & -\eta \\ -L & 0 & 1} + \bsmat{L & \rho^2 & -1 \\ -\mu & 0 & 1}^\top \bsmat{0 & \lambda_2 \\ \lambda_2 & 0} \bsmat{L & \rho^2 & -1 \\ -\mu & 0 & 1} \preceq \rho^2 \bmat{\iden , \zero }^\top P \bmat{\iden , \zero}.
\end{equation}
This is the LMI for minimizing a $\mu$-strongly convex $L$-smooth function (see~\cite{lessard2016analysis}), which has a better convergence rate compared to our minimax problem (i.e., any feasible $\rho$ of the LMI for 1-dimension minimax problem is also feasible for \eqref{eq:lmi-min}) because it is a special case of \eqref{eq:bilinear-sp} with $\bB = 0$ in the 1-dimensional case. So far, we show that as long as the LMI for the case of $m = n = 1$ holds, then the general case also holds since the general case can be decomposed into many 1-dimensional problems. This completes the proof.
\end{proof}

\section{Additional Results on SVHN}
\begin{figure}[h]
\centering
    \begin{subfigure}[b]{0.45\columnwidth}
        \centering
        \includegraphics[width=0.9\columnwidth]{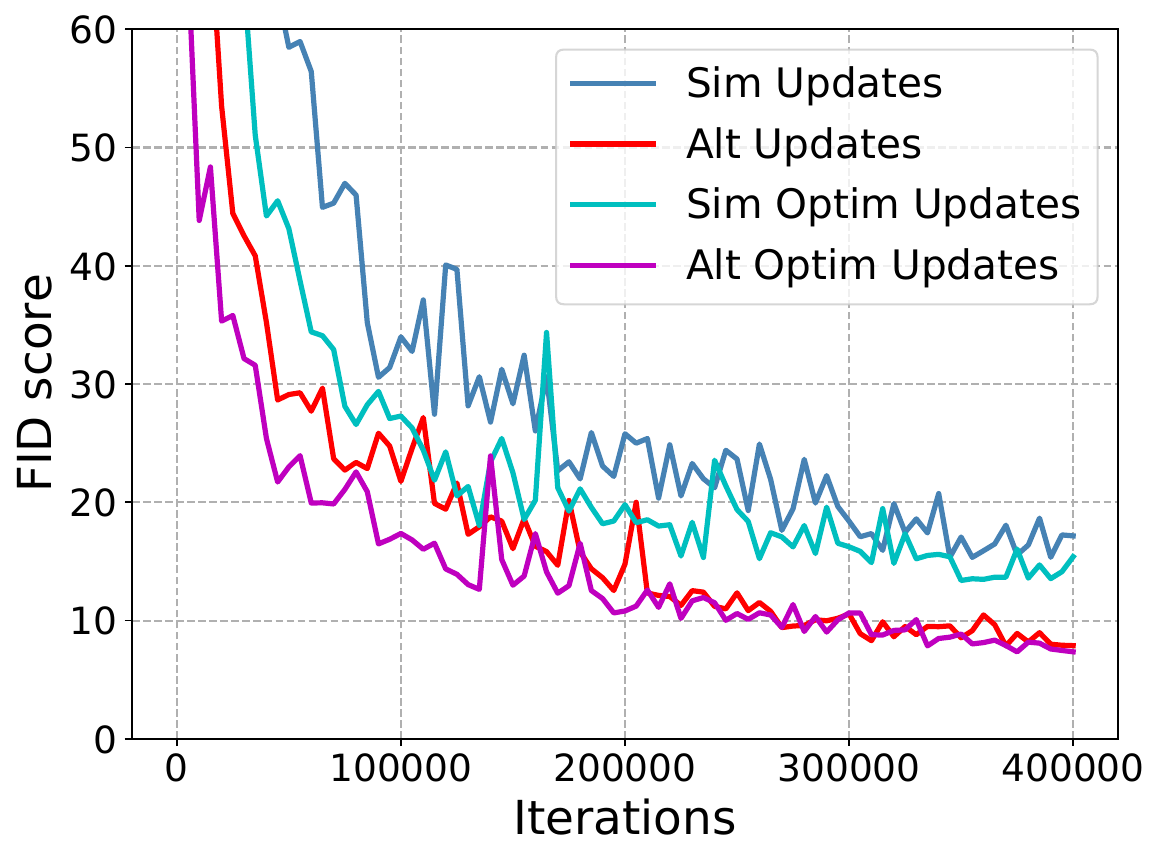}
        \vspace{-0.2cm}
        \caption{SGD}
	\end{subfigure}
	\begin{subfigure}[b]{0.45\columnwidth}
        \centering
        \includegraphics[width=0.9\columnwidth]{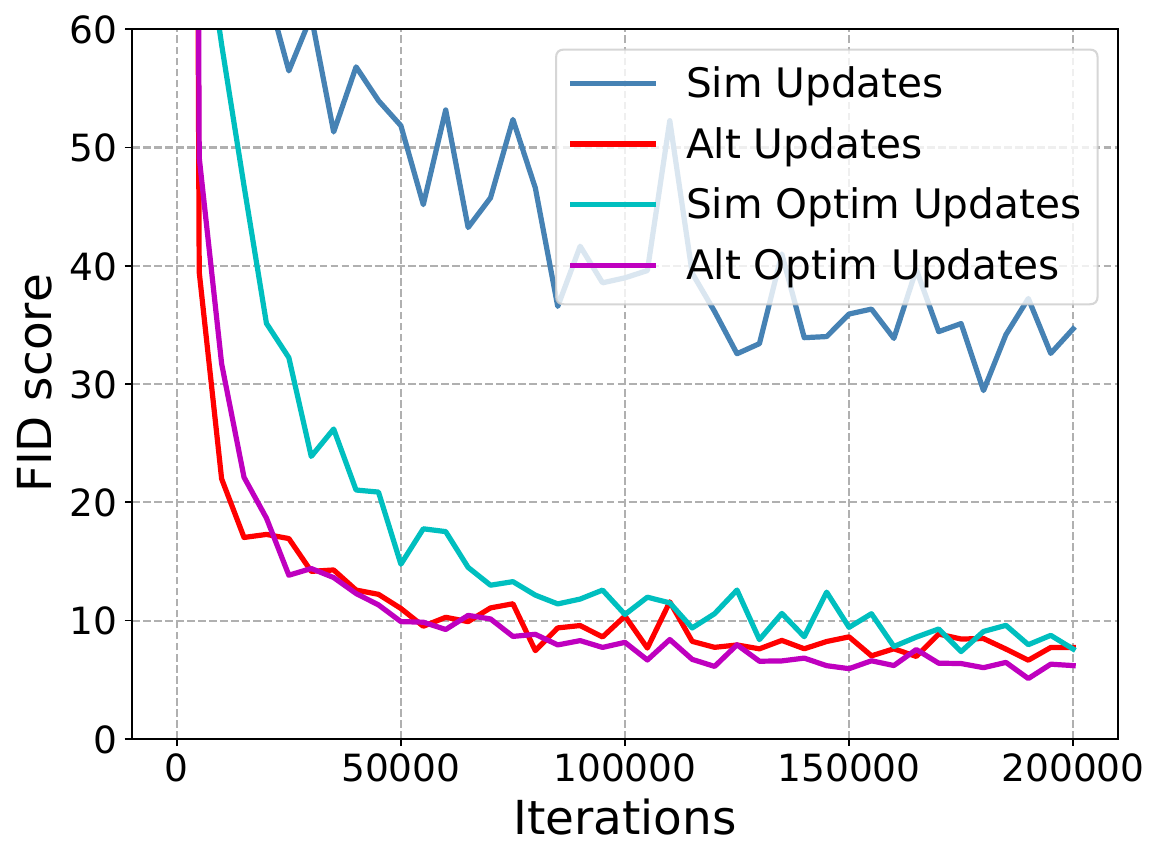}
        \vspace{-0.2cm}
        \caption{AMSGrad}
	\end{subfigure}
	\vspace{-0.2cm}
	\caption{\textbf{(a)} ResNet model trained with SGD on SVHN. \textbf{(b)} ResNet model trained with AMSGrad on SVHN. Alternating algorithms dominate simultaneous ones. Again, the use of optimism makes little difference for alternating algorithms.}
	\label{fig:resnet-gan-svhn}
\end{figure}

\section{Implementation Details for Generative Adversarial Networks}\label{app:imp}
For our experiments, we used the PyTorch\footnote{\url{https://pytorch.org/}} deep learning framework. For experiments, we compute the FID score using the provided implementation in Tensorflow\footnote{\url{https://github.com/bioinf-jku/TTUR}} for consistency with related works. 

\textbf{Optimistic update rule:} the simultaneous version of optimistic gradient descent-ascent takes the following form:
\begin{equation}
\begin{aligned}
    \bx_{t+1} &= \bx_t - 2\eta \nabla_\bx f(\bx_t, \by_t) + \eta \nabla_\bx f(\bx_{t-1}, \by_{t-1}) \\
    \by_{t+1} &= \by_t + 2\eta \nabla_\by f(\bx_t, \by_t) - \eta \nabla_\by f(\bx_{t-1}, \by_{t-1})
\end{aligned}
\end{equation}
By comparison, the alternating version iterates as follows:
\begin{equation}
\begin{aligned}
    \bx_{t+1} &= \bx_t - 2\eta \nabla_\bx f(\bx_t, \by_t) + \eta \nabla_\bx f(\bx_{t-1}, \by_{t-1}) \\
    \by_{t+1} &= \by_t + 2\eta \nabla_\by f(\bx_{t+1}, \by_t) - \eta \nabla_\by f(\bx_{t}, \by_{t-1})
\end{aligned}
\end{equation}

\textbf{Loss functions:} For DCGAN experiments, we used WGAN-GP objective~\citep{gulrajani2017improved}. For ResNet experiments, we used the hinge version of the adversarial non-saturating loss, see~\cite{miyato2018spectral}. As a reference, our ResNet architectures for CIFAR-10 and SVHN~\citep{netzer2011reading} have approximately $85$ layers in total for the generator and discriminator, including the nonlinearity and the normalization layers. This ResNet architecture was also used in~\cite{chavdarova2020taming}, see Appendix E 2.2 of~\cite{chavdarova2020taming}.

\textbf{Hyperparameters:} We conduct grid search over the step size (and $\beta_2$ for AMSGrad) for each setting. For SGD, the search range of step-size is $\{5\mathrm{e}{-4}, 1\mathrm{e}{-3}, 2\mathrm{e}{-3}, 5\mathrm{e}{-3}, 1\mathrm{e}{-2}, 2\mathrm{e}{-2}\}$. For AMSGrad, the search range of step-size is $\{5\mathrm{e}{-5}, 1\mathrm{e}{-4}, 2\mathrm{e}{-4}, 5\mathrm{e}{-4}, 1\mathrm{e}{-3}, 2\mathrm{e}{-3}\}$ while the search range of $\beta_2$ is $\{0.9, 0.99, 0.999\}$. We report the optimal hyperparameters used in the following tables. All these hyperparameters are tuned with random seed $1$. We have also tried other seeds (including seed $2$ and $3$) and the optimal hyperparameters could be different with different random seeds. However, the optimal curves across different random seeds look similar.
\begin{table}[h]
\caption{Hyperparameters for DCGAN experiments.}
\begin{center}
\begin{tabular}{ c|cccc } 
 Parameter & Sim-SGD & Alt-SGD & Sim-AMSGrad & Alt-AMSGrad \\ 
 \hline
 batch-size & 128 & 128 & 128 & 128 \\ 
 step-size (G) & 0.002 & 0.005 & 0.0005 & 0.0005 \\
 step-size (D) & 0.002 & 0.005 & 0.0005 & 0.0005 \\
 momentum & 0.0  & 0.0 & - & - \\
 $\beta_1$ & - & - & 0.0 & 0.0 \\
 $\beta_2$ & - & - & 0.999 & 0.999
\end{tabular}
\end{center}
\end{table}

\begin{table}[h]
\caption{Hyperparameters for ResNet experiments on CIFAR-10.}
\begin{center}
\resizebox{\textwidth}{!}{
\begin{tabular}{ c|cccc|cccc } 
 \multirow{2}{*}{Parameter} & \multicolumn{4}{c}{Simultaneous} & \multicolumn{4}{c}{Alternating}  \\ 
 & SGD & OSGD & AMSGrad & OAMSGrad & SGD & OSGD & AMSGrad & OAMSGrad \\
 \hline
 batch-size & 128 & 128 & 128 & 128 & 128 & 128 & 128 & 128 \\ 
 step-size (G) & 0.005 & 0.005 & 0.0002 & 0.0002 & 0.01 & 0.01 & 0.0005 & 0.001 \\
 step-size (D) & 0.005 & 0.005 & 0.0002 & 0.0002 & 0.01 & 0.01 & 0.0005 & 0.001 \\
 momentum & 0.0 & 0.0 & - & - & 0.0  & 0.0 & - & - \\
 $\beta_1$ & - & - & 0.0 & 0.0 & - & - & 0.0 & 0.0 \\
 $\beta_2$ & - & - & 0.999 & 0.99 & - & - & 0.999 & 0.999
\end{tabular}}
\end{center}
\end{table}

\begin{table}[h]
\caption{Hyperparameters for ResNet experiments on SVHN.}
\begin{center}
\resizebox{\textwidth}{!}{
\begin{tabular}{ c|cccc|cccc } 
 \multirow{2}{*}{Parameter} & \multicolumn{4}{c}{Simultaneous} & \multicolumn{4}{c}{Alternating}  \\ 
 & SGD & OSGD & AMSGrad & OAMSGrad & SGD & OSGD & AMSGrad & OAMSGrad \\
 \hline
 batch-size & 128 & 128 & 128 & 128 & 128 & 128 & 128 & 128 \\ 
 step-size (G) & 0.002 & 0.005 & 0.0001 & 0.0002 & 0.005 & 0.01 & 0.0005 & 0.0002 \\
 step-size (D) & 0.002 & 0.005 & 0.0001 & 0.0002 & 0.005 & 0.01 & 0.0005 & 0.0002 \\
 momentum & 0.0 & 0.0 & - & - & 0.0  & 0.0 & - & - \\
 $\beta_1$ & - & - & 0.0 & 0.0 & - & - & 0.0 & 0.0 \\
 $\beta_2$ & - & - & 0.999 & 0.999 & - & - & 0.999 & 0.99
\end{tabular}}
\end{center}
\end{table}

\end{document}